\documentclass[10pt,twocolumn,letterpaper]{article}

\usepackage{iccv}
\usepackage{times}
\usepackage{epsfig}
\usepackage{graphicx}
\usepackage{amsmath}
\usepackage{amssymb}
\usepackage{enumitem}
\usepackage{makecell}
\usepackage{color}
\usepackage[accsupp]{axessibility}

\usepackage[breaklinks=true,bookmarks=false]{hyperref}

\iccvfinalcopy 

\def\iccvPaperID{2278} 

\ificcvfinal\fi

\begin{document}

\title{Learning Data-Driven Vector-Quantized Degradation Model \\for Animation Video Super-Resolution}

\author{Zixi Tuo\thanks{This work was done while Zixi Tuo was a research intern at Microsoft Research Asia.} \textsuperscript{1}, 
Huan Yang\thanks{Corresponding author.} \textsuperscript{2}, 
Jianlong Fu\textsuperscript{2}, 
Yujie Dun\textsuperscript{1}, Xueming Qian\textsuperscript{1,3}\\
\textsuperscript{1}Xi’an Jiaotong University \quad\textsuperscript{2}Microsoft Research Asia\\ \textsuperscript{3}Shaanxi Yulan Jiuzhou Intelligent Optoelectronic Technology Co., Ltd\\
{\tt\small zixit99@gmail.com, \{huayan, jianf\}@microsoft.com, \{dunyj, qianxm\}@mail.xjtu.edu.cn}
}

\maketitle
\ificcvfinal\thispagestyle{empty}\fi

\begin{abstract}
   Existing real-world video super-resolution (VSR) methods focus on designing a general degradation pipeline for open-domain videos while ignoring data intrinsic characteristics which strongly limit their performance when applying to some specific domains (\eg, animation videos). In this paper, we thoroughly explore the characteristics of animation videos and leverage the rich priors in real-world animation data for a more practical animation VSR model. In particular, we propose a multi-scale \textbf{V}ector-\textbf{Q}uantized \textbf{D}egradation model for animation video \textbf{S}uper-\textbf{R}esolution (VQD-SR) to decompose the local details from global structures and transfer the degradation priors in real-world animation videos to a learned vector-quantized codebook for degradation modeling. A rich-content \textbf{R}eal \textbf{A}nimation \textbf{L}ow-quality (RAL) video dataset is collected for extracting the priors. We further propose a data enhancement strategy for high-resolution (HR) training videos based on our observation that existing HR videos are mostly collected from the Web which contains conspicuous compression artifacts. The proposed strategy is valid to lift the upper bound of animation VSR performance, regardless of the specific VSR model. Experimental results demonstrate the superiority of the proposed VQD-SR over state-of-the-art methods, through extensive quantitative and qualitative evaluations of the latest animation video super-resolution benchmark. The code and pre-trained models can be downloaded at \href{https://github.com/researchmm/VQD-SR}{https://github.com/researchmm/VQD-SR}.
\end{abstract}

\begin{figure}[t]
\begin{center}
\includegraphics[width=\linewidth]{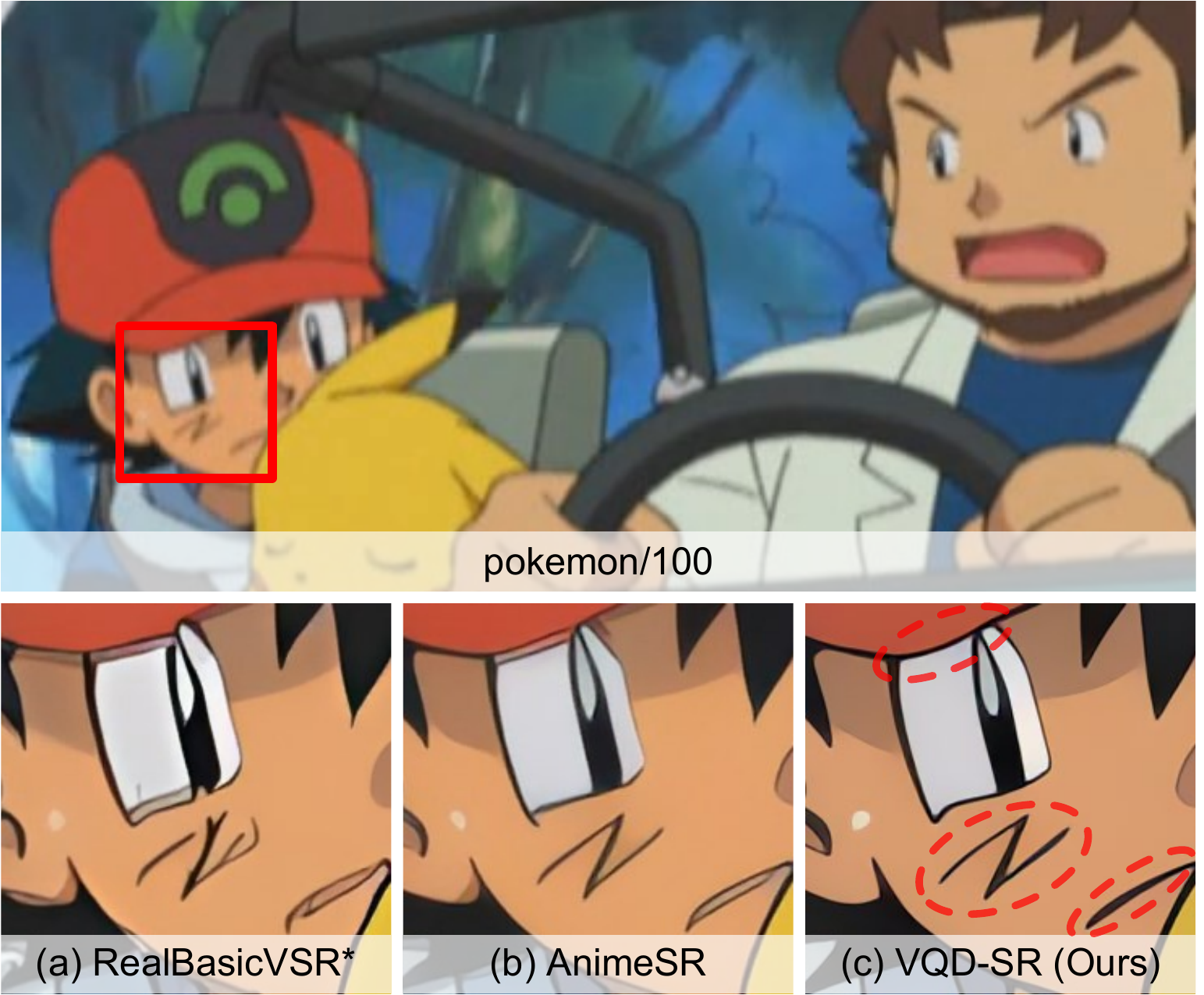}
\end{center}
   \caption{\textbf{A comparison between our proposed VQD-SR and other SOTA VSR methods in real scenario.} As indicated by the red dotted circles, our VQD-SR could restore more appealing details by considering the degradation priors in real animation videos. (`$*$' denotes fine-tune on animation videos).}
\label{fig:teaser}
\end{figure}
\vspace{-0.6cm}

\section{Introduction}
Animation video super-resolution (VSR)~\cite{AnimeSR}, which is a subdiscipline of video super-resolution (VSR)~\cite{chan2021basicvsr,chan2022basicvsr++,RealBasicVSR,liu2022learning,tian2020tdan,xie2022mitigating}, aims to restore high-resolution (HR) videos from their low-resolution (LR) counterparts in animation domain. As a special type of visual art, animation video shows great entertainment, cultural, educational and commercial values. However, most of them available on the Web are in limited resolutions and have visually unappealing artifacts due to the passing ages, lossy compression and transmission. Dedicated to real applications, animation VSR is a blind SR~\cite{luo2022learning,wang2021real,zhang2021designing,yang2022degradation,yin2023online} task which is more challenging than conventional VSR tasks~\cite{chan2021basicvsr,chan2022basicvsr++,isobe2020video,liu2022learning,tian2020tdan,xiao2023online}, because of the complicated and agnostic real-world degradations. Thus, practical VSR methods for improving real-world LR animation videos are in highly demanded.

Blind SR model for open-domain has long been the focus. As it's hard to collect LR-HR pairs in real scenarios, recent learning-based approaches try to model the real-world degradations from HR to LR, thus generating pseudo LR-HR pairs to train SR models. These approaches can be roughly divided into two categories, explicit modeling and implicit modeling, with regard to the degradation procedures. Explicit modeling~\cite{gu2019blind,liu2020learning,wang2021real,zhang2021designing} methods try to combine basic degradation operators like noise, blur and rescaling to imitate the real-world degradation pipeline. Instead, implicit modeling methods employ a neural network, which is usually in the manner of generative adversarial networks (GAN)~\cite{goodfellow2020generative}, as an alternative to the combination of basic degradation operators. However, these general degradation methods designed for open-domain videos are not quite suitable for the animation domain as they ignore the intrinsic characteristics of the data. Directly adopting existing degradation models for open-domain to animation VSR would lead to unpleasant results (as shown in Fig.~\ref{fig:teaser} a), even though they are fine-tuned on the animation videos.

Specific to animation VSR, AnimeSR~\cite{AnimeSR} moves one step further to propose a learnable basic operator (LBO) with an ``input-rescaling'' strategy that considers the data characteristics and demonstrates their advantages over open-domain VSR methods. However, the proposed ``input-rescaling'' strategy of constructing pseudo HR-LR pairs for the LBO training requires human annotations, which prevents their method from scaling up to modeling various real-world degradations. In AnimeSR~\cite{AnimeSR}, only three LBOs are trained with three human-annotated videos respectively, leading to unappealing results (as shown in Fig.~\ref{fig:teaser} b). Thus, how to effectively leverage large-scale real-world animation videos based on their data characteristics is still an open problem.

In this paper, we propose a multi-scale \textbf{V}ector-\textbf{Q}uantized \textbf{D}egradation model for animation video \textbf{S}uper-\textbf{R}esolution (VQD-SR), which leverages the data characteristics of animation videos. We observe that different from open-domain videos with complex textures and irregular illumination conditions, animation videos are roughly composed of basic visual patterns like \textit{smooth color patches and clear lines} which can be easily captured with high fidelity by vector-quantization (VQ)~\cite{VQGAN,razavi2019generating,van2017neural}. Such data characteristics could also be used for encoding and transferring degradation priors from low-quality animation videos and results in a degradation codebook. Given any clean LR animation frame, the degraded LR counterpart can be easily obtained by looking up the pre-trained degradation codebook. Specifically, a novel multi-scale VQGAN for degradation modeling is proposed to learn and transfer real-world degradation priors from a large-scale \textbf{R}eal \textbf{A}nimation \textbf{L}ow-quality (RAL) video dataset collected from the Web. To improve the generalization ability of VQD-SR, we propose a two-stage training pipeline and a stochastic top-k VQ strategy to match up with the multi-scale VQGAN, yielding visually appealing results (as shown in Fig.~\ref{fig:teaser} c).

To further lift the upper bound of existing animation VSR methods, we propose an HR-SR strategy to enhance the quality of HR training videos for more ideal ground truths. This is motivated by our observations that existing HR animation videos are mostly collected from the Web which contains conspicuous compression artifacts and such artifacts could be easily reduced by existing SR methods. Our contributions are summarized as follows:
\begin{itemize}[nosep]
  \item We collect a large-scale real LR animation dataset, and conduct a comprehensive study on real animation data to thoroughly explore the challenges and data characteristics in animation VSR task.
  \item We propose a novel multi-scale VQ degradation model to transfer real-world degradation priors for synthesizing pseudo LR-HR training pairs. We further propose an HR-SR data enhancement strategy to improve the performance of existing VSR methods for animation.
  \item Extensive experiments demonstrate the effectiveness of our proposed methods which significantly improve the recent SOTA animation VSR method in both quantitative and qualitative comparisons.
\end{itemize}

\section{Related Work}
\subsection{Video Super-Resolution}
Recent years have witnessed a prosperity in deep learning based VSR methods~\cite{chan2022basicvsr++,RealBasicVSR,isobe2020video,qiu2022learning,tian2020tdan,wang2019edvr,xu2021temporal}. Different from single-image super-resolution (SISR), the auxiliary temporal information is available in VSR for the restoration of inner-frame features. Sliding-window structure~\cite{tian2020tdan,wang2019edvr,xu2021temporal,yi2019progressive} and recurrent structure~\cite{chan2021basicvsr,chan2022basicvsr++,fuoli2019efficient,haris2019recurrent,isobe2020video} are two common paradigms in VSR to utilize the inter-frame information. However, it also faces challenges while taking the advantages. The degradations could be more complicated in videos and may accumulate along the time sequence~\cite{RealBasicVSR,xie2022mitigating}, causing the performance drop in real-world scenarios. 

There have been several attempts in blind-SR field to deal with real-world degradations, which could be categorized into explicit and implicit modeling. Explicit methods model the degradation as the effects of objective operators (or kernels). Methods like ~\cite{gu2019blind,zhang2018learning} assume these degradations to be simple combinations of blur, noise and down-sampling kernels, therefore conducting conditional restoration based on the prediction of kernel parameters. Operators like JPEG/FFmpeg~\cite{liu2020learning} for lossy compression, $sinc$ filter~\cite{wang2021real} for ringing artifacts and strategies like high-order modeling~\cite{wang2021real}, random shuffling~\cite{zhang2021designing}, are further investigated in latest studies to simulate the real degradation pipeline and expand the synthesis space. However, these methods merely depend on predefined operators, and ignore to utilize the information contained in real data, thus there's still a gap between the synthesis LR and real LR data. Implicit methods~\cite{luo2022learning,maeda2020unpaired,wei2021unsupervised,yuan2018unsupervised} discard the objective degradation operators, but learn the distribution of real data, therefore realize the degradation by converting the HR input to fit the real LR distribution. Yet, the learned distribution space is uncontrollable, thus the degradation network may easily project unseen out-of-distribution samples to unpredictable suboptimal space causing annoying artifacts.

\subsection{Animation Video Processing}
Animation is a type of visual art that is represented in a nonrealistic artistic style. A diversity of animation processing technologies based on deep learning have been developed these years, such as animation style transfer~\cite{chen2020animegan,hong2021domain,wang2022fine}, animation video colorization~\cite{lei2019fully,thasarathan2019automatic,zhang2019deep}, animation video generation~\cite{gupta2018imagine} and animation video interpolation~\cite{siyao2021deep}. AnimeSR~\cite{AnimeSR} is a recent study for real-world animation VSR tasks, which develops LBO to expand the basic degradation operators. In this paper, we take a further step to explore the characteristics of animation videos, and make special designs in consideration of these characteristics to improve the performance of real-world animation VSR.

\subsection{Vector-Quantized Codebook}
Vector-quantized(VQ) codebook was first proposed in VQ-VAE~\cite{van2017neural} to learn the discrete representation for images in latent space. Later, a hierarchical VQ-VAE-2~\cite{razavi2019generating} is further proposed. To improve the perceptual quality, a discriminator and perceptual loss were added in VQGAN~\cite{VQGAN}. Recently in VQFR~\cite{VQFR}, a VQ-based face restoration method was also proposed. Although initially designed to avoid the``posterior collapse'' issue of VAE or tokenize input images for a more effective transformer modeling, we discover that the idea of VQ is naturally suitable for representing animation video frames and the form of codebook is capable to internalize the real degradation priors. 

\section{Methodology}
In this section, we first discuss the characteristics of real animation videos and the challenges for animation VSR task in Sec.~\ref{sec:real LQ}. Based on the observations in Sec.~\ref{sec:real LQ}, we introduce the proposed real-world animation degradation model in Sec.~\ref{sec:degradation model}. Finally, we illustrate our HR training data enhancement strategy in Sec.~\ref{sec:gt enhance}.

\subsection{Real-World Degradation in Animation Videos}
\label{sec:real LQ}
To thoroughly understand the characteristics of animation videos and the challenges in animation VSR task, we collect a \textbf{R}eal \textbf{A}nimation \textbf{L}ow-quality (RAL) video dataset. RAL contains over 10K LR frames extracted from 441 real-world low-quality animation videos. To the best of our knowledge, RAL is the largest real-world low-quality video dataset in the animation domain. Example frames are shown in Fig.~\ref{fig:RAL} (statistics of RAL can be found in the supplementary). To construct such a dataset, we first download numerous earlier animation videos from multiple video sites such as YouTube and Bilibili. Then we manually select videos according to their resolutions, quality, contents and styles. To ensure the dataset contains rich animation structures and degradation priors, we adopt a simple scene filtering procedure to remove the frames which have subtle motions by evaluating their average absolute differences, thus reducing the redundancy.

Animation videos are roughly composed of \textit{smooth color patches and clear lines}, while the latter ones usually play an important role in portraying objects and expressing visual semantics. In contrast to the fact that degradation losses for open domain videos are mainly about high-frequency textures, degradations for animation videos mostly happen around the edges which are more sensitive according to the human perceptual. Such degradations usually act as intermittent edges, aliasing edges, ringing artifacts, and rainbow effects, as shown in Fig.~\ref{fig:RAL}. These observations of the proposed RAL dataset inspire us that the key challenge in animation VSR is to restore visually natural and clear edges.

Degradation modeling has been the focus of many previous blind-SR works. Because of the obvious differences between animation and natural style videos, degradation methods designed for natural videos cannot properly handle degradations in animation videos. As a result, these methods suffer from amplified artifacts causing phenomena like perturbation edges, bleeding edges and inaccurate colorization along edges, even though they are fine-tuned on the animation videos (as shown in Fig.~\ref{fig:teaser} a).

\begin{figure}[t]
\begin{center}
\includegraphics[width=\linewidth]{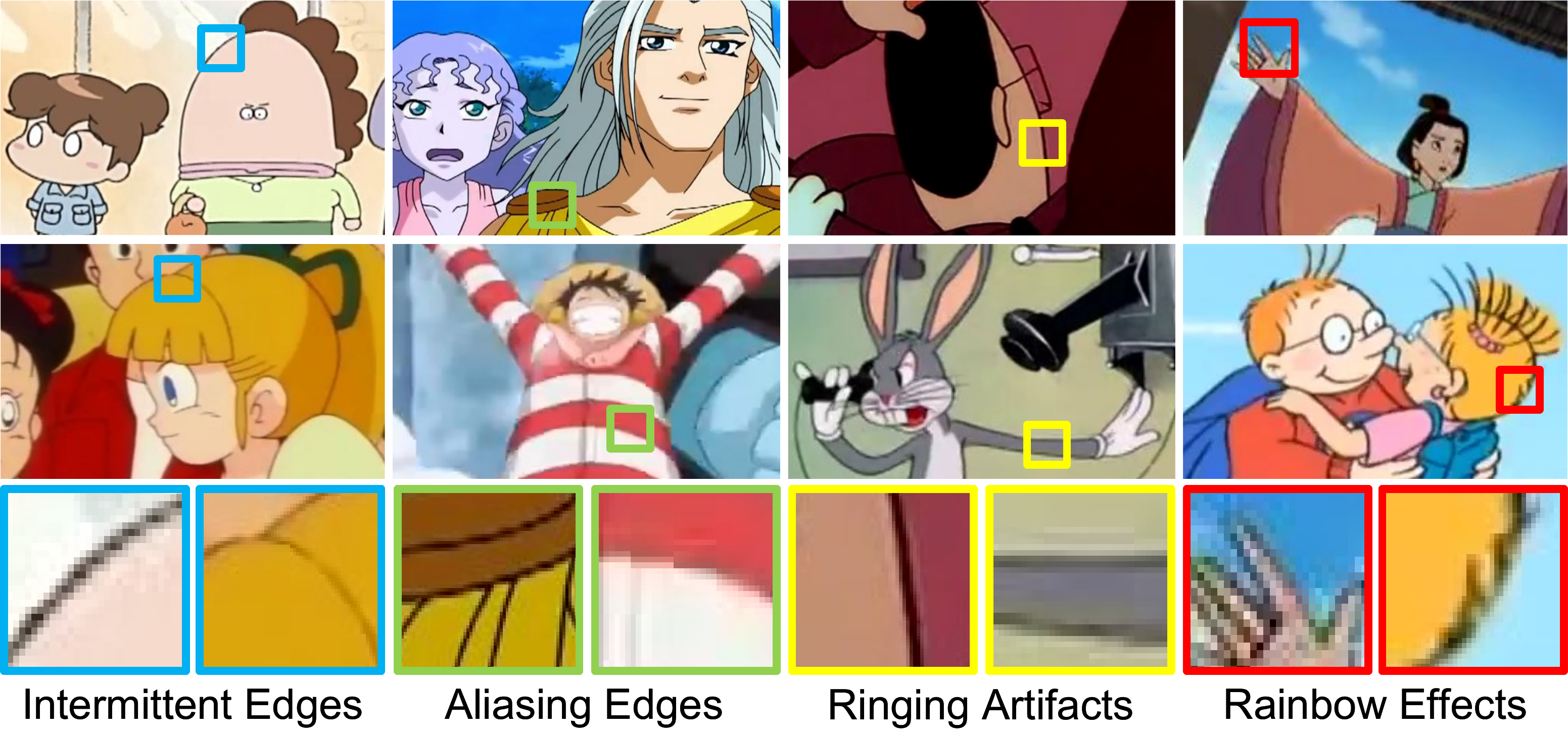}
\end{center}
   \caption{Samples of RAL and typical degradation phenomena in real-world LR animation videos. RAL is a real-world animation video dataset contains over 10K LR frames. Intermittent edges, aliasing edges, ringing artifacts and rainbow effects are four typical degradation phenomena in real-world animation videos.}
\label{fig:RAL}
\end{figure}

\begin{figure*}[t]
\begin{center}
   \includegraphics[width=\linewidth]{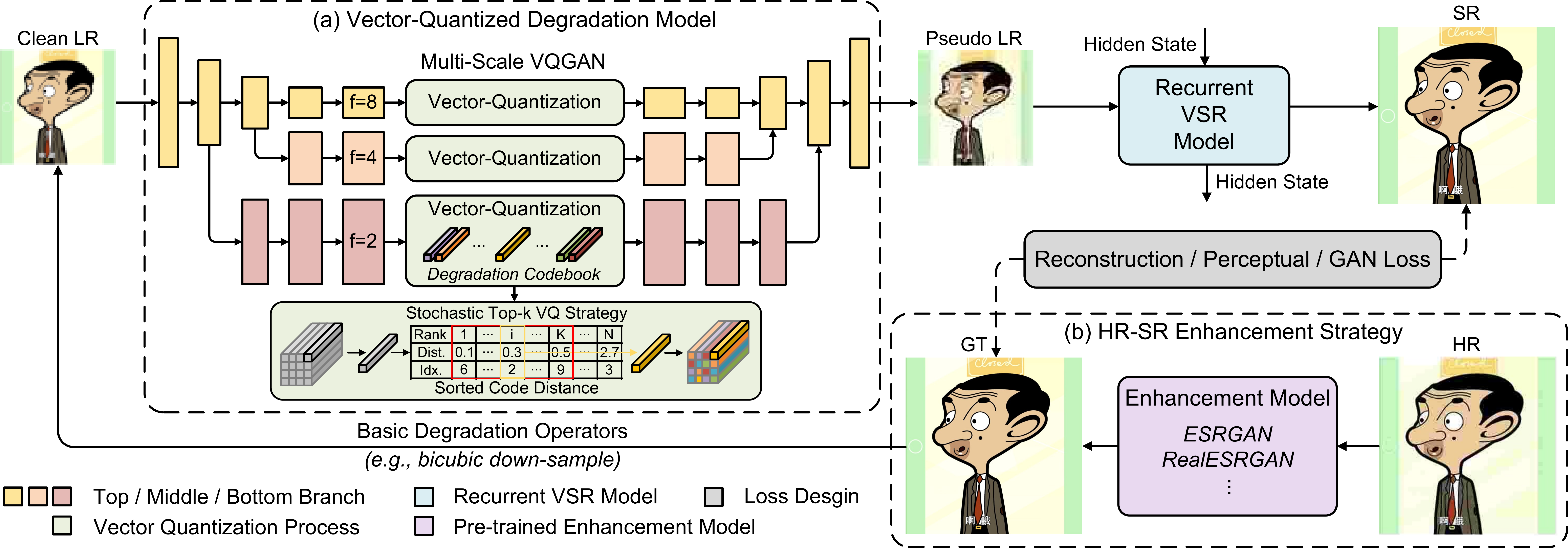}
\end{center}
   \caption{The overview of VQD-SR for training the animation VSR model. (a) Vector-quantized degradation model is a multi-scale VQGAN which transfers the real-world degradations from the learned VQ codebook to clean LR video frames to synthesize pseudo LR videos for VSR training, with the stochastic top-k VQ strategy. (b) HR-SR enhancement strategy eliminates the artifacts in HR animation training videos with an existing SR model to generate more ideal ground truths for supervising the VSR results.}
\label{fig:model}
\end{figure*}

\subsection{Vector-Quantized Degradation Modeling}
\label{sec:degradation model}
Instead of imitating the real-world degradation pipeline by simply combining and adjusting some predefined basic operators~\cite{wang2021real,zhang2021designing}, we propose to explore the data intrinsic characteristics further and utilize degradation priors of real-world animation videos. As illustrated in Sec.~\ref{sec:real LQ}, animation videos are roughly composed of \textit{smooth color patches and clear lines}. Compared to open-domain videos, which contain complex textures and irregular illumination conditions, it is possible to encode such animation video frames into a codebook of finite basic visual patterns (i.e., color patches and lines). More importantly, any animation video frame can be further decoded with corresponding patterns with high fidelity. If such a process is restricted to the real LR domain (\eg, the proposed RAL dataset), the degradation priors of real animation videos will be encapsulated into the aforementioned codebook simultaneously and could be further used for synthesizing pseudo LR data for the VSR training.

VQGAN~\cite{VQGAN} is a reconstruction network composed of an encoder $E$, a decoder $G$ and a vector-quantized codebook $Z$. Motivated by the perceptually rich codebook, we resort to VQGAN~\cite{VQGAN}, trained on the proposed RAL dataset with rich degradations of animation videos, to realize the decomposition of low-quality frames and encapsulate degradation priors into a learned codebook. When given a clean LR frame, the real-world degradation priors can be transferred by replacing the original clean patches with the corresponding degraded versions by looking up the learned degradation codebook. In this section, we first introduce our modified multi-scale VQGAN, then illustrate the two-stage training pipeline, and finally use the trained model to synthesize pseudo LR-HR pairs in diverse levels of degradations by a stochastic top-k vector-quantization strategy.

\noindent
\textbf{Multi-Scale VQGAN.} 
As discussed in Sec.~\ref{sec:real LQ}, the degradations in animation videos mostly happen around the edges, which are fine-grained local details. A degradation model should capable of processing these local details while keeping the global structures (\eg, colors and shapes) at the same time. However, these two goals are usually conflicted with each other under the scope of the original single-scale VQGAN~\cite{VQGAN}, as these two kinds of components are entangled together. Disentangling local details from global structures is a requisite for degradation modeling with VQGAN.

The multi-scale design has been widely explored in low-level vision~\cite{mei2020image, yang2020learning} and achieved great success, especially in the image restoration task. Inspired by this, we propose a multi-scale VQGAN which contains three parallel encoding branches. As shown in Fig.~\ref{fig:model} a, these three branches share a shallow feature extraction module but have independent deeper encoding modules with different compression factors $f$ (we choose $f=\{8,4,2\}$ for the top, middle and bottom branches respectively). Benefiting from our multi-scale design, image information of different levels could be effectively disentangled and learned by each encoding branch in one VQ-GAN model.

\noindent
\textbf{Two-Stage Training Pipeline.}
However, it is non-trivial to achieve the feature disentanglement as we expected by directly training the proposed model from scratch. This is because the bottom branch with the smallest scale will dominate the model training since it provides the easiest way to model converge. We call this phenomenon ``short-cut''. Such a problem prevents the top and the middle branches from learning image information for different scales, and the learned image representations are still entangled with each other. To take full advantage of our multi-scale VQGAN, we propose a two-stage training pipeline. We first remove the middle and the bottom branches, and only train the top branch with the largest compression factor $f$. In this stage, the top branch is forced to take effect and be responsible to reconstruct the global structures. After the convergence of the top branch, we add the other two branches back to fine-tune the overall multi-scale VQGAN. In the second stage, the two new branches only need to recover the remaining smaller structures and fine-grained details, thus avoiding the ``short-cut'' problem. 

\noindent
\textbf{Stochastic Top-k VQ Strategy.}
After training our multi-scale VQGAN on a real-world low-quality animation video dataset (\eg, the proposed RAL dataset in Sec.~\ref{sec:real LQ}), the learned codebooks will contain rich degradation information and are capable of transferring such degradations into unseen animation data. Specifically, as shown in Fig.~\ref{fig:model}, the learned multi-scale VQGAN will be used as the degradation model to synthesize pseudo LR images from clean LR images for further VSR model training. Although the degradation priors in the codebooks are diverse, there is only one fixed degradation output for each input as the quantization process takes the nearest neighbor search strategy to reconstruct the input, leading to poor generalization capability of the final VSR model. To expand the degradation space and introduce more randomness during training, we adopt a stochastic top-k search strategy when looking up the codebook for the VQ process. More precisely, given the encoded output $\hat{z} = E(x) \in\mathbb{R}^{h \times w \times n_z}$ from the input image $x$, the former vector quantization methods search the nearest entry in the codebook $\mathcal{Z}=\{z_l\}^L_{l=1}$ for each vector $\hat{z}_{ij}$ in position $(i,j)$ to get the quantized output $z_q$:
\begin{equation}
    z_q = q(\hat{z}) := \big(\arg \mathop{\min}\limits_
    {z_l \in\mathcal{Z}}
    \|\hat{z}_{ij}-z_l\| \big) \in\mathbb{R}^{h \times w \times n_z}. \\
\end{equation}

We expand the nearest search strategy to a stochastic top-k search strategy. That is, we obtain the $k_{th}$ closest entry in the codebook for each vector $\hat{z}_{ij}$ in position $(i,j)$:
\begin{equation}
    z_q = q(\hat{z}) := \big(\arg \mathop{k_{th}}\limits_
    {z_l \in\mathcal{Z}}
    \|\hat{z}_{ij}-z_l\| \big) \in\mathbb{R}^{h \times w \times n_z}. \\
\end{equation}
For each sample in every training iteration, the value of $k$ is randomly chosen from a predefined range $[1, K]$ and controls the level of degradation. We chose K=50 in VQD-SR and the experiment of the choice of K can be found in the supplementary.

A key point here is that we adopt the stochastic top-k strategy only for the bottom branch while keeping the nearest neighbor search for the top and middle branches. This is because the larger-scale structures like colors and shapes are encoded in these two branches, and we do not want them to be changed when augmenting the local details for degradation. That is also the reason why we make a great effort to design the multi-scale VQGAN that encodes the different components in input images separately. The idea of stochastic top-k VQ strategy also corresponds with the fact that one HR image could have multiple degraded versions.

\subsection{HR-SR Enhancement Strategy}
\label{sec:gt enhance}
AVC~\cite{AnimeSR} is the latest large-scale high-quality animation video dataset, specified for animation VSR tasks. As opposed to images that could be recorded in lossless format, most of the animation videos available on the Web all went through compression to reduce the network traffic. As demonstrated in Fig.~\ref{fig:model} b, the HR animation video frames still suffered from the compression artifacts, and therefore are not completely ideal as the training ground truths. Based on this observation, we suggest enhancing the HR animation videos to further lift the upper bound for the existing animation VSR model. Since the contents of animation videos are relatively simple and the HR video frames are relatively clean, general pre-trained SR models (\eg, Real-ESRGAN~\cite{wang2021real}) could be used to reduce the artifacts while not contaminating image details. We thus enhance all the HR video frames in the AVC dataset by first applying an SR model, then downsampling to align the original sizes.

\subsection{Training Details}
\noindent
\textbf{Degradation Model.}
In our multi-scale VQGAN, we set the compression factor $f$ to \{8,4,2\} for the \{top, middle, bottom\} branch respectively and use 1024 codebook entries with 256 channels in total. The input images are randomly cropped to patches of $ 256 \times 256$. The training of our multi-scale VQGAN is a two-stage procedure. We follow~\cite{VQGAN} to adopt Adam~\cite{kingma2014adam} optimizer with the base learning rate of $ 4.5 \times 10^{-6} $ for both stages. The learning rate in top branch is decreased by 4 times in the second stage. We set the total batch size to 32 and 24 for the two stages, respectively. We only apply VQ loss $ L_{vq} $~\cite{van2017neural} and perceptual loss $ L_{per}$~\cite{zhang2018unreasonable} in the first stage, while adding the adversarial loss $ L_{adv}$~\cite{VQGAN} in the second stage. The full objective function for encoder $E$, decoder $G$ and hierarchical VQ codebook $Z$ is:
\begin{equation}
    L\big( E,G,Z \big) =  L_{vq} + L_{per} + L_{adv}, \\
\end{equation}
in which $ L_{vq}$ is calculate by:
\begin{equation}
\begin{split}
     L_{vq} = & \| x-\hat{x} \| + \beta \| sg[E(x)_{(s)}]-z_{q(s)} \|_2^2 \\
     &+ \| sg[z_{q(s)}]-E(x)_{(s)} \|_2^2 .
\end{split}
\end{equation}
Where $x,\hat{x}$ are the input and reconstructed images, $sg[\cdot]$ denotes the stop-gradient operation, and $ s \in \{ t,m,b\}$ means the top, middle or bottom branches. More implementation details can be found in the supplementary.

\noindent
\textbf{Video Super-Resolution Model.} We adopt the unidirectional recurrent network in AnimeSR~\cite{AnimeSR} as the VSR model because of its efficiency, and further remove the SR feedback in the recurrent block to ensure a more stable training according to our observation that too many recurrent features would sometimes cause a collapse of GAN loss as the omitting of explicit alignment module. We train the VSR model in two stages. In the first stage, we pre-train the VSR model with $L1$ loss for 300K iterations, with batch size 16 and learning rate $10^{-4}$. In the second stage, we further add perceptual loss~\cite{johnson2016perceptual} and GAN loss~\cite{ledig2017photo} to fine-tune the network for another 300K iterations, with batch size 16 and learning rate $5\times10^{-5}$. As the promotion of perceptual quality is mainly during the second stage, we only append our multi-scale VQGAN for more realistic degradations in the second stage to reduce the training cost while only adopting basic operators in the first stage.

\section{Experiments}
\subsection{Datasets and Metrics}
We evaluate our method and compare its performance with other SOTA SR methods on AVC~\cite{AnimeSR}, the latest proposed large-scale animation video clip dataset. AVC is composed of three subsets. The training set AVC-Train contains 553 high-quality clips, and the testing set AVC-Test contains 30 high-quality clips. There is also a AVC-RealLQ , which contains 46 real-world low-quality clips for evaluation. Each clip consists of 100 frames. We adopt MANIQA~\cite{yang2022maniqa}, the winner in NTIRE 2022 NR-IQA challenge~\cite{gu2022ntire}, as the no-reference image quality evaluation metric, as it in the previous animation VSR work~\cite{AnimeSR}.

\subsection{Comparisons with State-of-the-art Methods}
We compare our VQD-SR with six SOTA SR methods. Among them, AnimeSR~\cite{AnimeSR} and RealBasicVSR~\cite{RealBasicVSR} are blind video super-resolution (VSR) methods. Real-ESRGAN~\cite{wang2021real} and BSRGAN~\cite{zhang2021designing} are blind single-image super-resolution (SISR) methods with explicit degradation modeling. We also conduct experiments on BasicVSR~\cite{chan2021basicvsr} and PDM~\cite{luo2022learning} as two representative methods for none-blind VSR method and blind SISR with implicit degradation modeling. Apart from AnimeSR~\cite{AnimeSR}, for which we report the result in the original paper, all the others are general SR methods for open-domain images or videos. For fair comparisons, we fine-tune their officially released models on animation dataset AVC-Train.

\noindent
\textbf{Quantitative Comparison.}
As shown in Tab.~\ref{tab:maintable}, we evaluate all the methods on AVC-RealLQ~\cite{AnimeSR} for quantitative comparisons. Among them, Real-ESRGAN~\cite{wang2021real} and BSRGAN~\cite{zhang2021designing} expand the former explicit degradation models by introducing high-order degradations and random shuffling respectively, which greatly improve the synthetic ability. And with the help of powerful SR backbone (RRDBNet~\cite{wang2018esrgan} with 16.70M parameters), their results are quite remarkable. Nonetheless, ignoring the intrinsic characteristics of animation data limits their performance when applied to animation domain. Although specialized for animation videos, AnimeSR~\cite{AnimeSR} only utilizes a small number of real data (three human-annotated animation videos), which hinders the performance of VSR model in real scenarios. Different from them, our VQD-SR considers the intrinsic characteristics of animation videos and leverages the enormous degradation priors contained in rich-content real animation videos. VQD-SR also adopts the HR enhancement strategy for more effective SR supervision. Due to these advantages, VQD-SR achieves a result of 0.4096 and significantly outperforms the SOTA animation VSR model~\cite{AnimeSR} by 0.0264 in MANIQA on AVC-RealLQ~\cite{AnimeSR}. Meanwhile, as the removal of SR recurrent feedback in VSR model, the number of parameters in VQD-SR are further reduced to 1.47 M with the runtime of 25 ms computed on a NVIDIA Tesla M40 GPU with the LR size of $180 \times 320$, which demonstrates the superiority of our VQD-SR in real applications. 

\begin{table}[]
    \centering
    \caption{Quantitative comparison on AVC-RealLQ for $4 \times$ animation VSR. `$*$' denotes fine-tune on animation dataset AVC-Train~\cite{AnimeSR}. \textbf{Bold} text indicates the best performance.\\}
    \label{tab:maintable}
    \small
    \resizebox{\linewidth}{!}{
    \begin{tabular}{l|c|c|c}
        \Xhline{1.2pt}
        Method & Params (M) & Runtime (ms) & MANIQA$\uparrow$ \\
        \Xhline{0.4pt}
        BasicVSR*~\cite{chan2021basicvsr} & 6.29 & 175 & 0.2829 \\
        PDM*~\cite{luo2022learning} & 16.70 & 438 & 0.2574 \\
        Real-ESRGAN*~\cite{wang2021real} & 16.70 & 438 & 0.3825 \\
        BSRGAN*~\cite{zhang2021designing} & 16.70 & 438 & 0.3836 \\
        RealBasicVSR*\cite{RealBasicVSR} & 6.29 & 252 & 0.3504 \\
        \Xhline{0.4pt}
        AnimeSR~\cite{AnimeSR} & 1.50 & 26 & 0.3832 \\
        VQD-SR (Ours) & \textbf{1.47} & \textbf{25} & 
        \textbf{0.4096} \\
        \Xhline{1.2pt}
    \end{tabular}}
\end{table}

\begin{figure}[t]
\begin{center}
\includegraphics[width=\linewidth]{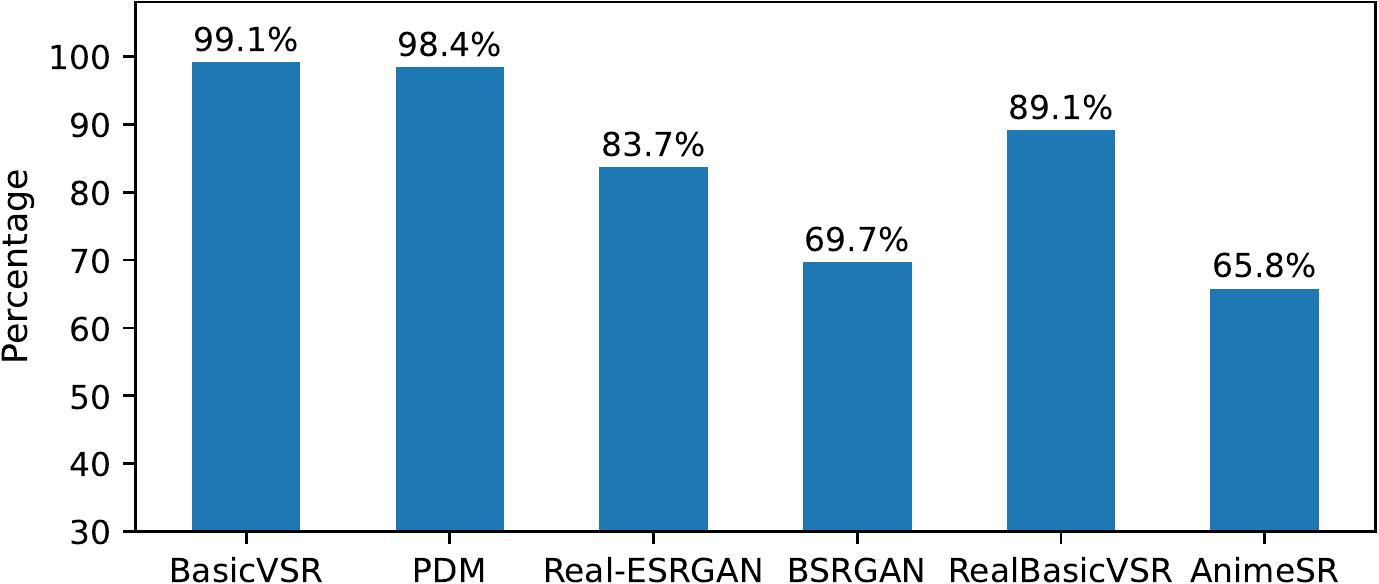}
\end{center}
   \caption{Results of User Study. The values on Y-axis represent the percentage of users that prefer VQD-SR over other approaches. }
\label{fig:user_study}
\end{figure}

\begin{figure*}[h]
\begin{center}
   \includegraphics[width=\linewidth]{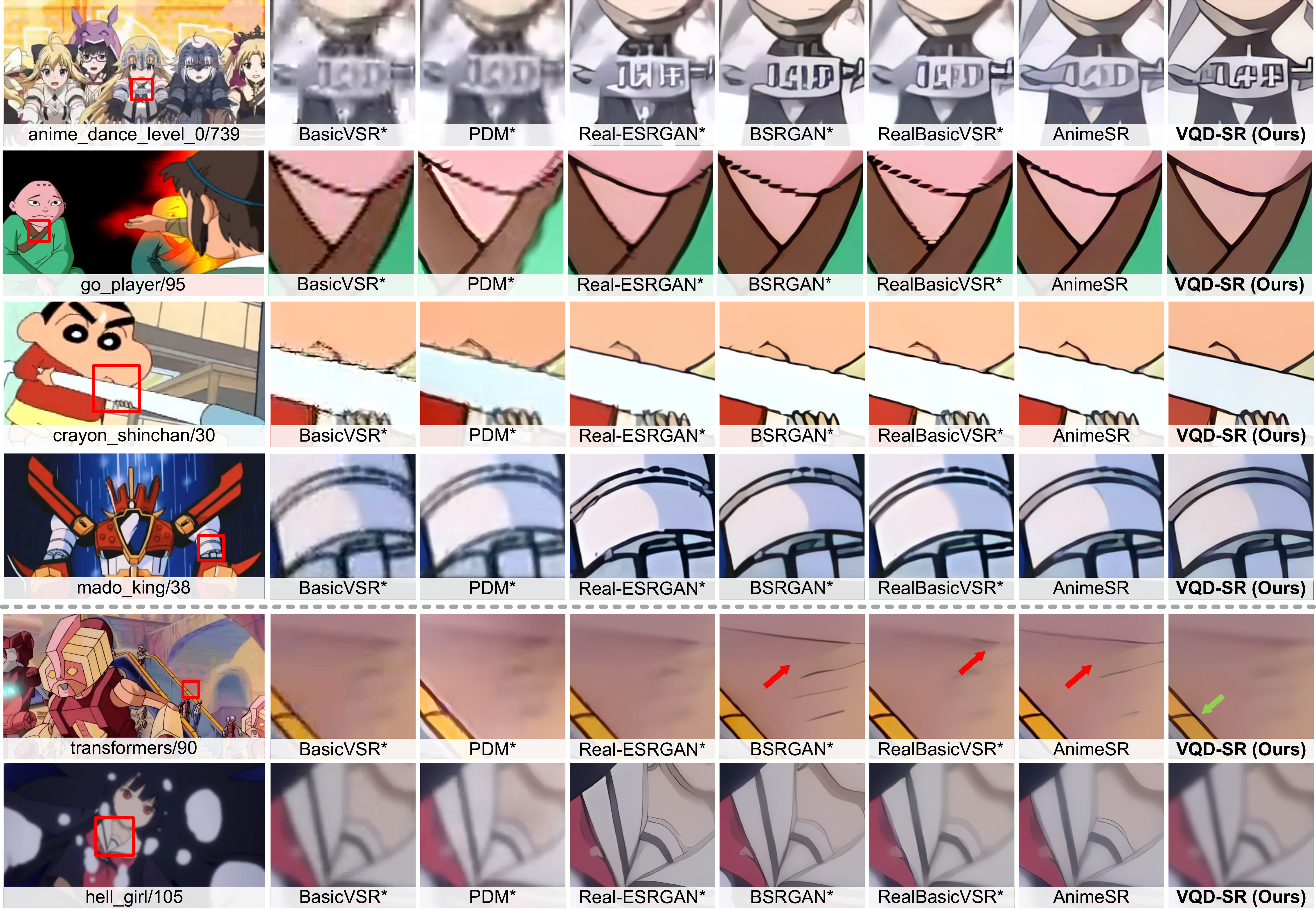}
   \caption{Qualitative comparisons on AVC-RealLQ~\cite{AnimeSR} for $4\times$ scaling factor. `$*$' denotes fine-tune on animation domain dataset. The frame name is shown at the bottom of each case. Our VQD-SR recovers visually natural and sharper lines with fewer artifacts (the first 4 rows) and is also capable to handle some intended scenarios with fewer over-sharp artifacts (row 5th with smooth \textcolor{red}{background} and clear \textcolor{green}{foreground}, row 6th in underwater scene). Zoom in to see better visualization.}
   \label{fig:main_result}
\end{center}
\end{figure*}

\noindent
\textbf{Qualitative Comparison.}
As shown in Fig.~\ref{fig:main_result}, we compare visual qualities of different approaches on real-world AVC-RealLQ video clips. It can be observed that VQD-SR greatly improves the visual quality, especially for the recovery of clear lines. Additionally, the results of VQD-SR are more visually natural with fewer artifacts in some intended scenarios (\eg, the out-focus background blur and smooth underwater scene in row 5-6),  while most of the other methods suffer from over-sharp artifacts like irregular lines and patterns. More visual comparisons with original inputs for reference and result analysis are in the supplementary materials. 

We also conduct a user study to further evaluate the visual quality of our approach. There are 20 subjects involved in this user study and 22080 votes are collected on AVC-RealLQ~\cite{AnimeSR} with 46 video clips. For each comparison, we adopt an A-B test that provides the users with two images in random order which include one VQD-SR frame. Users are asked to select the one with higher visual quality. As shown in Fig.~\ref{fig:user_study}, comparing with the SR methods designed for open-domain, over $80\%$ of users voted for our VQD-SR. By considering the data characteristic of animation, there are still over $65\%$ users who prefer our method to AnimeSR~\cite{AnimeSR}. 

\subsection{Ablation study}
\label{sec:ablation}

In this section, we conduct ablation studies on each significant component of our VQ degradation model and HR-SR enhancement strategy in VQD-SR.

\begin{figure}[h]
\begin{center}
\includegraphics[width=\linewidth]{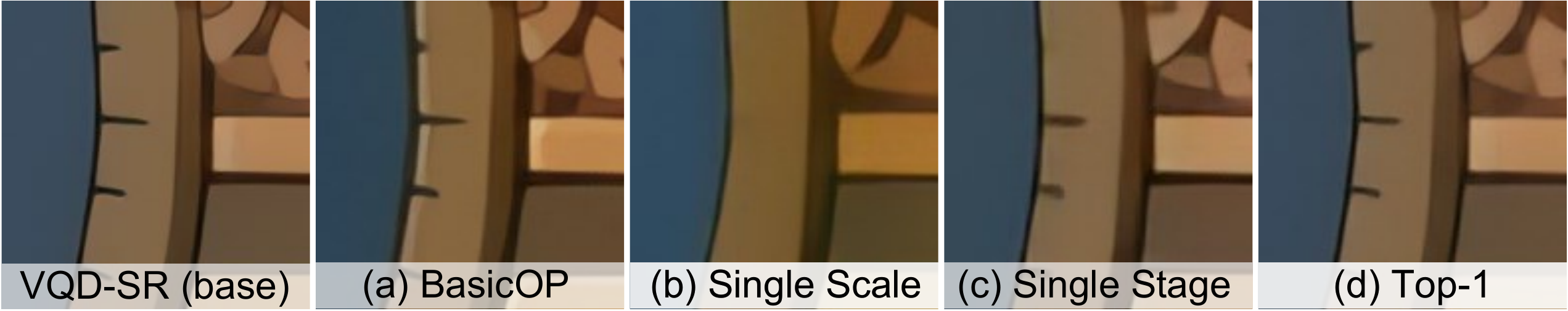}
\end{center}
   \caption{Ablation study on VQ degradation model.} 
\label{fig:lq_ablation}
\end{figure}

\noindent
\textbf{VQ Degradation Model.} In the ablation study of VQ degradation model design, we conduct all the experiments based on VQD-SR (base), which denotes a plain version of VQD-SR trained without HR-SR enhancement. As shown in Tab.~\ref{tab:vq_ablation} a, we first compare the degradation model which is simply composed of basic operators (blur, noise and FFmpeg) to our VQD-SR (base). With the help of real-world degradation priors contained in VQ codebook, the result can be improved from 0.3768 to 0.3857 in MANIQA. Refering to Fig.~\ref{fig:lq_ablation} a, VSR model trained with only basic operators fails to handle some real-world degradations, leading to artifacts around the edges. We further compare the structure and training strategy of VQ degradation model in Tab.~\ref{tab:vq_ablation} b and Tab.~\ref{tab:vq_ablation} c. As the global structures and local details are entangled together in single-scale VQ model and the multi-scale model trained in single-stage, the performances dropped by 0.0391 and 0.0427, respectively. As shown in Fig.~\ref{fig:lq_ablation} b and Fig.~\ref{fig:lq_ablation} c, some fine-grained details are missing. For VQ strategies, as shown in Tab.~\ref{tab:vq_ablation} d and Fig.~\ref{fig:lq_ablation} d, our stochastic top-k VQ strategy can improve MANIQA by 0.0087 compared with the original nearest neighbor search strategy, as the top-k VQ strategy enables multi-level degradations which improves the generalization ability of the trained animation model, leading to more appealing results.

\begin{figure}[]
\begin{center}
\includegraphics[width=\linewidth]{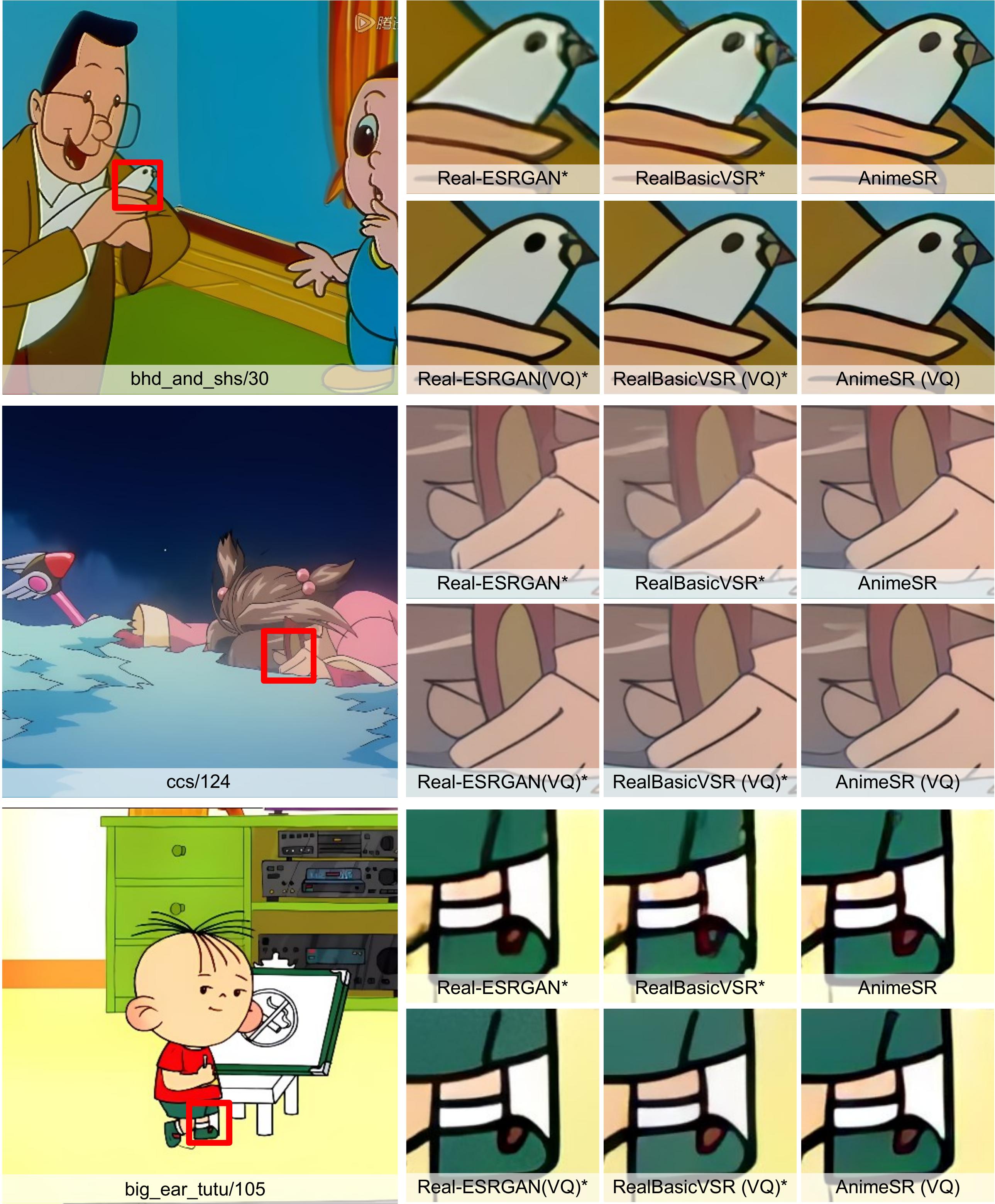}
\end{center}
   \caption{Applying VQ degradation to widely used SR backbones. Settings are consistent apart from degradation methods. Models trained with VQ degradation could produce more clear and vi-
sual natural lines with fewer artifacts.  \\ }
\label{fig:vqd_ablation}
\end{figure}

To further illustrate the general applicability, we also apply our VQ degradation model to other SOTA SR backbones, Real-ESRGAN and RealBasicVSR, fine-tuned with animation data in the same setting. Referring to Tab.~\ref{tab:vq_applicable}, SR models with VQ degradation achieve better results comparing with their original degradation methods, which indicates that the proposed VQ degradation is generally effective for animation SR models. It is noteworthy that the performance gains could be more significant for larger backbones. The visual examples in Fig.~\ref{fig:vqd_ablation} also show the superiority of our VQ degradation method, especially to recover clear and visual natural lines which are the most import patterns deciding the visual quality of animation videos.
\begin{table}[]
    \centering
    \caption{Ablation study results of degradation models, structures of VQ degradation model, training pipelines of VQ degradation model and VQ codebook search strategies. The MANIQA scores are reported for comparisons.\\}
    \label{tab:vq_ablation}
    \small
    \begin{tabular}{c|c|c}
        \Xhline{1.2pt}
        & \multicolumn{2}{c}{\textbf{(a) Degradation Model}} \\
        \Xcline{2-3}{0.4pt}
         MANIQA$\uparrow$   &   BasicOP-Only   &   VQD-SR (base)  \\
        \Xcline{2-3}{0.4pt}
                             &     0.3768       &      0.3857   \\ 
         \Xhline{0.4pt}
         
        &\multicolumn{2}{c}{\textbf{(b) VQGAN Structure}} \\
        \Xcline{2-3}{0.4pt}
          MANIQA$\uparrow$&     Single-scale     &     VQD-SR (base)\\
        \Xcline{2-3}{0.4pt}
                          &      0.3466     &      0.3857    \\
         \Xhline{0.4pt}
        
        & \multicolumn{2}{c}{\textbf{(c) Training Strategy}}\\
        \Xcline{2-3}{0.4pt}
         MANIQA$\uparrow$ & Single-stage  & VQD-SR (base)  \\
        \Xcline{2-3}{0.4pt}
        &   0.3430   & 0.3857  \\
        \Xhline{0.4pt}
        
        & \multicolumn{2}{c}{\textbf{(d) Codebook Search}}\\
        \Xcline{2-3}{0.4pt}
        MANIQA$\uparrow$ &   Nearest     &   VQD-SR (base)   \\
        \Xcline{2-3}{0.4pt}
                         &   0.3770   &   0.3857 \\
        
        \Xhline{1.2pt}
    \end{tabular}
\end{table}

\begin{table}[]
    \centering
    \caption{Applying VQ degradation to widely used SR backbones. For each backbone, settings are consistent apart from degradation methods.}
    \label{tab:vq_applicable}
    \resizebox{\linewidth}{!}{
    \begin{tabular}{c| c c c}
    \Xhline{1.2pt}
    Backbone & Real-ESRGAN* & RealBasicVSR* & AnimeSR \\
    \Xhline{0.4pt}
    Params (M) & 16.7 & 6.3 & 1.5 \\
    Original degradation & 0.3825  & 0.3504 & 0.3832 \\
    VQ degradation & \textbf{0.3967} & \textbf{0.3849} & \textbf{0.3857} \\
    \Xhline{1.2pt}
    \end{tabular}
    }
\end{table}

\begin{figure}[]
\begin{center}
\includegraphics[width=\linewidth]{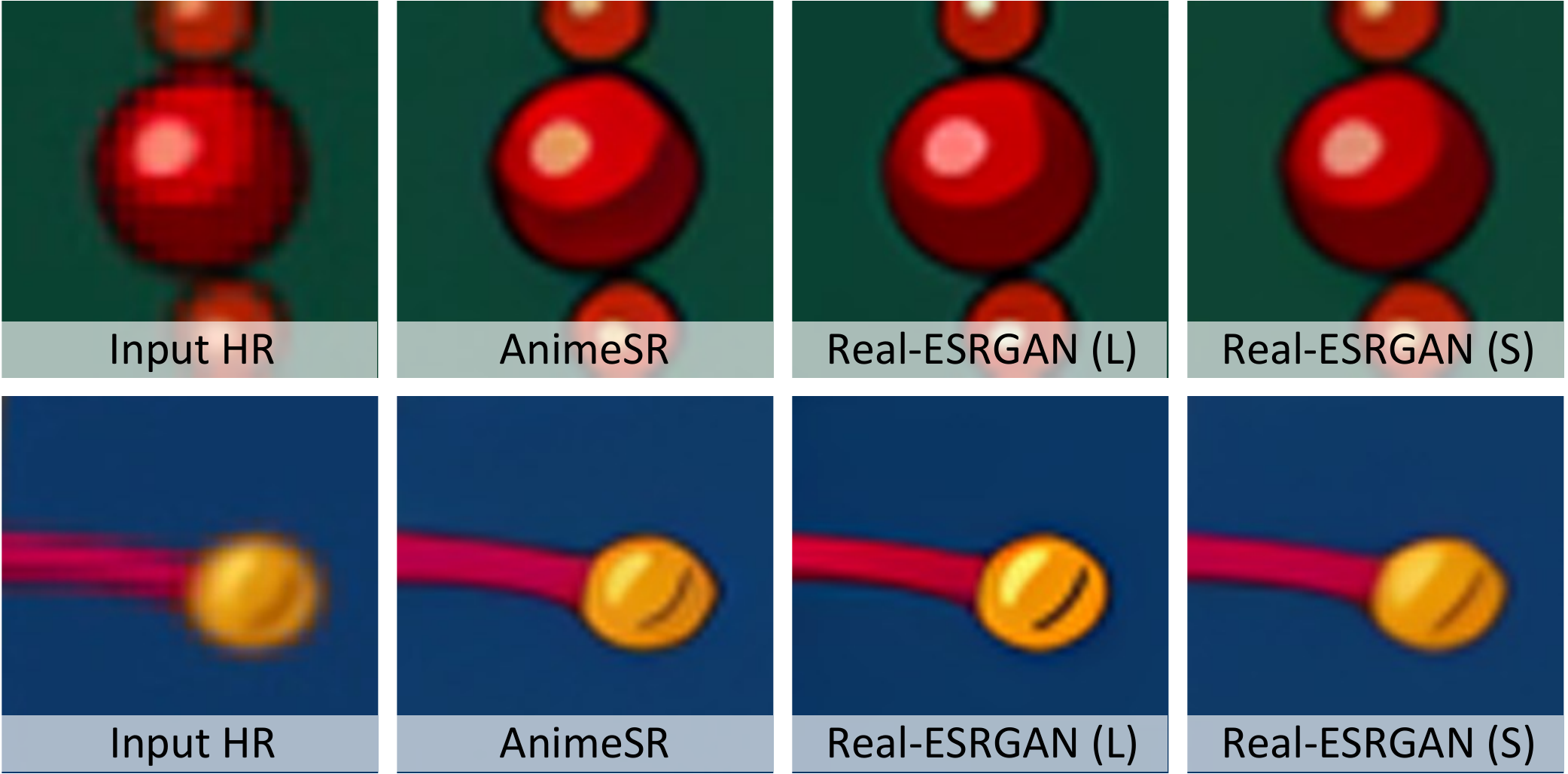}
\end{center}
   \caption{Enhanced HR data with HR-SR strategy by different enhancement models. \\ }
\label{fig:hr_ablation_sota}
\end{figure}

\noindent
\textbf{HR Enhancement.}
The training HR animation videos collected from the Web also contain artifacts which are not ideal for ground truths, hindering the performances of animation VSR models. We propose the HR-SR strategy to enhance the quality of HR animation videos for more effective training ground truths. 
We first compare three different HR enhancement models by experiments on VQD-SR (base). The quantitative results are shown in Tab.~\ref{tab:modelforhr-sr}. All three models improve the performance of animation VSR model, which verifies that the proposed HR-SR strategy is suitable for the training of animation VSR model, and improvements are not subject to the specific model for enhancement. However, as the SR process for data enhancement is conducted in HR domain, which takes a high demand on computing resources, we finally choose SISR model, Real-ESRGAN (S)~\cite{wang2021real} with relatively small model size. Visual examples of the enhanced HR frames can be found in Fig.~\ref{fig:hr_ablation_sota}.

\begin{figure}[]
\begin{center}
\includegraphics[width=\linewidth]{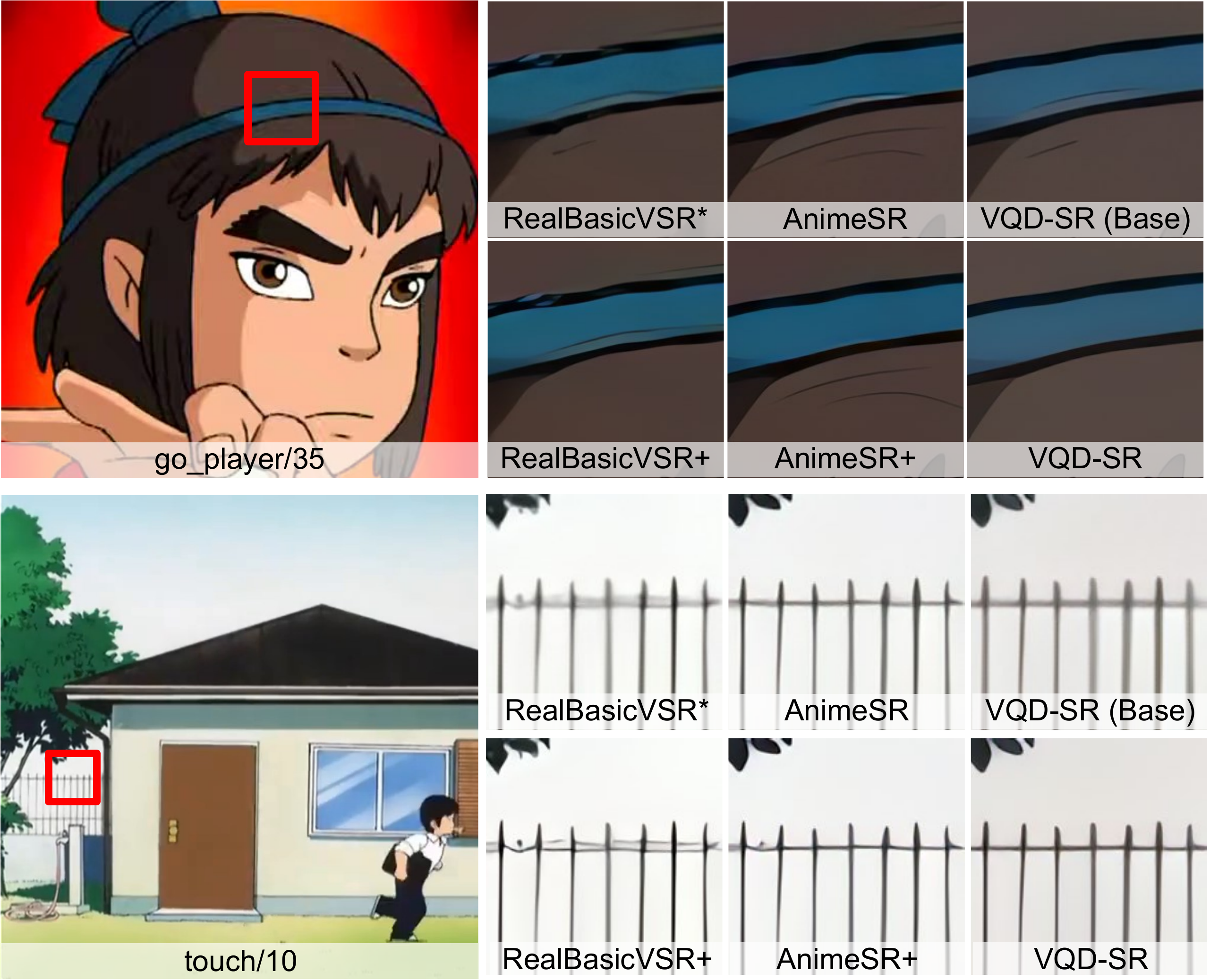}
\end{center}
   \caption{Ablation study of HR-SR enhancement strategy for different VSR models. HR-SR enhancement is generally effective task, regardless of the specific VSR model.}
\label{fig:hr_ablation_model}
\end{figure}

We further utilize the enhanced HR animation videos processed by Real-ESRGAN(S)~\cite{wang2021real} to fine-tune different VSR models. Apart from the proposed VQD-SR, RealBasicVSR~\cite{RealBasicVSR} is a blind VSR model designed for open-domain videos and AnimeSR~\cite{AnimeSR} is an VSR model specified for animation domain. As shown in Tab.~\ref{tab: results of hr-sr enhance}, with the help of HR enhancement, all three VSR models get improved, which illustrates that the proposed HR enhancement strategy is effective for animation VSR task, regardless of the specific VSR model. The visual results of RealBasicVSR+, AnimeSR+, and VQD-SR in Fig.~\ref{fig:hr_ablation_model} also verify the advantages of the HR-SR enhancement strategy for animation VSR.

\begin{table}[]
    \centering
    \caption{Ablation study of different SR models for HR-SR enhancement. All three models improve the performance. We choose Real-ESRGAN (S) considering the computing cost.\\}
    \label{tab:modelforhr-sr}
    \begin{tabular}{c|c}
        \Xhline{1.2pt}
                  Enhancement Method            & MANIQA$\uparrow$ \\
         \Xhline{0.4pt}
         No Enhancement &  0.3857 \\
         AnimeSR~\cite{AnimeSR}&  0.4146 \\
         Real-ESRGAN (L)~\cite{wang2021real}& 0.4211  \\
         Real-ESRGAN (S)~\cite{wang2021real} &    0.4096     \\
        \Xhline{1.2pt}
    \end{tabular}
\end{table}

\begin{table}[]
    \centering
    \caption{Ablation study of HR-SR enhancement for different animation VSR models on AVC-RealLQ. We report MANIQA for comparison.\\}
    \label{tab: results of hr-sr enhance}
    \small
    \begin{tabular}{c|cc}
        \Xhline{1.2pt}
            MANIQA$\uparrow$   &wo $/$HR-SR  &   w $/$HR-SR \\
         \Xhline{0.4pt}
         RealBasicVSR~\cite{RealBasicVSR}   & 0.3504    &     0.3855  \\
         AnimeSR ~\cite{AnimeSR}       & 0.3832    &  0.4079 \\
         VQD-SR  & 0.3857    &   0.4096   \\
        \Xhline{1.2pt}
    \end{tabular}
\end{table}

\section{Limitation and Discussion}
\label{sec:fail}
\begin{figure}[h]
    \centering
    \includegraphics[width=\linewidth]{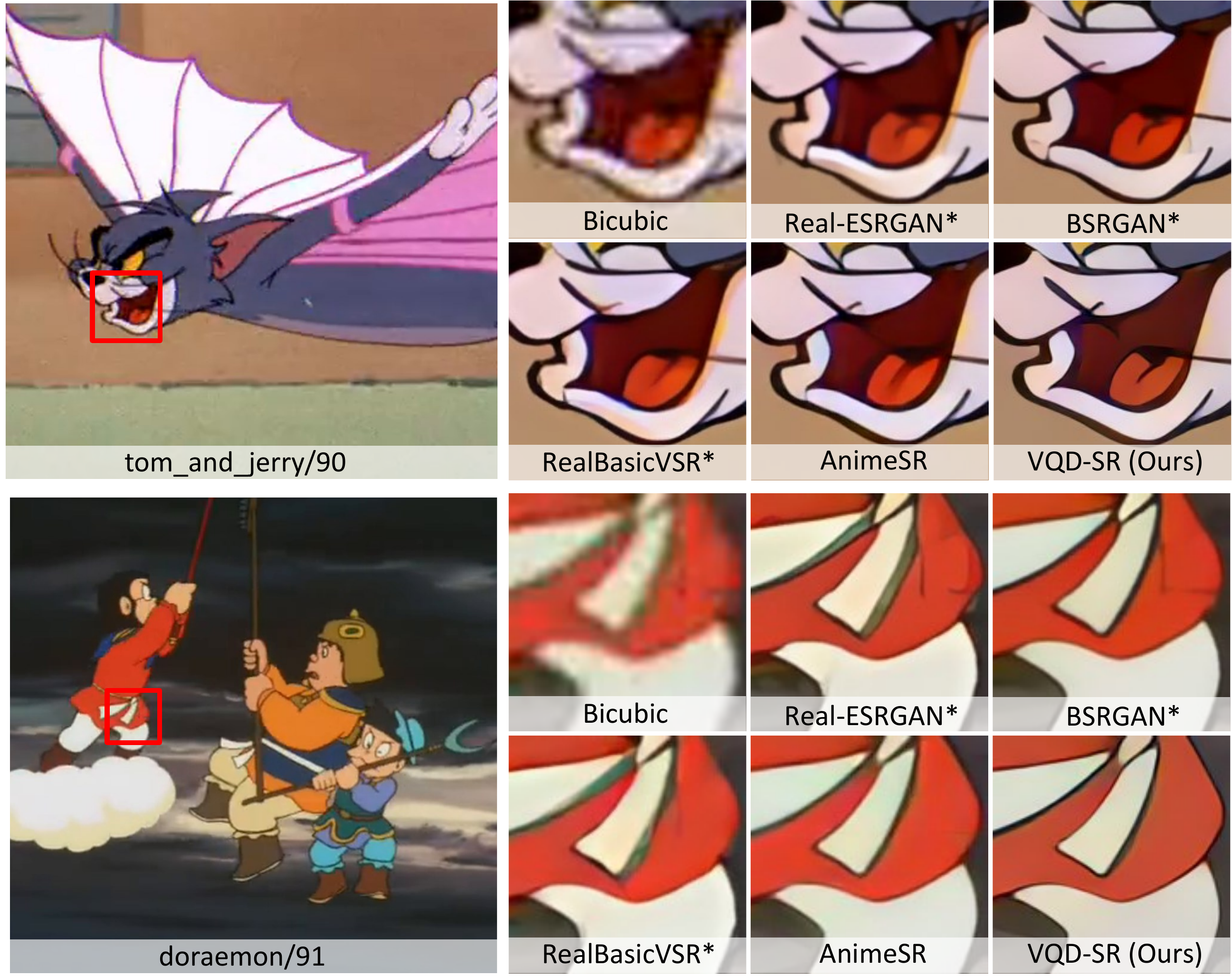}
    \caption{Failure cases when LR frames endure severe color distortions around edges.}
    \label{fig:sup_fail}

\end{figure}

In this section, we visualize the failure cases of VQD-SR in Fig.~\ref{fig:sup_fail}. Our VQD-SR is capable to restore sharper lines and suppress diverse artifacts in animation videos. However, in some extreme circumstances (\eg, severe lossy compressions, long passing ages), there still gets room for improvement, for example the remaining color distortions around edges. Nevertheless, comparing with other SOTA methods, our VQD-SR still achieves better performances. 

\section{Conclusion}
In this paper, we thoroughly study the characteristics of animation videos and fully leverage the rich prior knowledge contained in real data for a more practical animation VSR model. To be specific, we propose a novel degradation model which utilize a multi-scale VQGAN, self-supervised on purely real low-quality animation data (RAL), to decompose animation frames and establish a codebook containing diverse real-world degradation priors. We also propose a two-stage training pipeline and a stochastic top-k VQ strategy to match up with the VQ degradation model. We further propose to enhance the HR training videos to lift the performance ceiling for animation VSR tasks. Experiment results show the superiority of the proposed VQD-SR over existing SOTA models, as our methods could generate more effective LR-HR training pairs in view of the characteristics in real animation videos. 

\textbf{Acknowledgement.} This work was supported in part by the NSFC under Grant 62272380 and 62103317, the Science and Technology Program of Xi’an, China under Grant 21RGZN0017, and Microsoft Research Asia.


{\small
\bibliographystyle{ieee_fullname}
\bibliography{egbib}
}

\clearpage
\begin{appendix}

\section*{Supplementary Material}
In this supplementary material, Sec.~\ref{sec:inply} first illustrates the implementation details of VQD-SR, which includes the implementation of VQ degradation model in Sec.~\ref{subsec:degradation_model}, the details of VSR model in Sec.~\ref{subsec:vsr}, and the experiment details in Sec.~\ref{subsec:exper}. Then Sec.~\ref{sec:ral_sta} introduces our RAL dataset with statistics and representative samples. Sec.~\ref{sec:more_results} shows more comparison results.

\section{Implementation Details}
\label{sec:inply}
\subsection{VQ Degradation Model}
\label{subsec:degradation_model}
\noindent
\textbf{Network Architecture.} The architecture of our multi-scale VQGAN for degradation modeling is described in Tab.~\ref{tab:top_branch_arichtecture}. The inputs of (b) Middle Branch and (c) Bottom Branch are the intermediate outputs of (a) Top Branch, represented by $x_m$ and $x_b$ respectively. The corresponding output features $\hat{x}_m$ and $\hat{x}_b$ are further added back to the top branch when decoding. The vector-quantization is conducted pixel by pixel on the encoded outputs $z_t$, $z_m$ and $z_b$ in latent space with 256 channels VQ codebook sized 1024. The compression factor $f$, which denotes the patch size when projecting each entry in the VQ codebook from latent space to the original image space, is set to $\{$8,4,2$\}$ controlled by the number of downsample steps in these three branches. During training, the middle branch and the bottom branch only take effect in the second-stage. All the training procedures are performed on eight NVIDIA 32G V100 GPUs.

\noindent
\textbf{Degradation Pipeline.} The whole degradation pipeline with multi-scale VQGAN to transfer the real-world degradation priors can be formulated as: $x = D^n(y) = ($FFmpeg $\circ$ VQD $\circ$ Down $\circ$ Noise $\circ$ Blur$)(y)$. Where $x$ denotes the degraded LR clips and $y$ denotes the HR clips. For basic operators (blur, noise, and FFmpeg), we follow the settings and hyperparameters in AnimeSR~\cite{AnimeSR}. For VQ degradation, we keep the nearest neighbor search on the top and the middle branch while adopting the stochastic top-k VQ strategy on the bottom branch. 

\begin{figure}[]
    \centering
    \includegraphics[width=\linewidth]{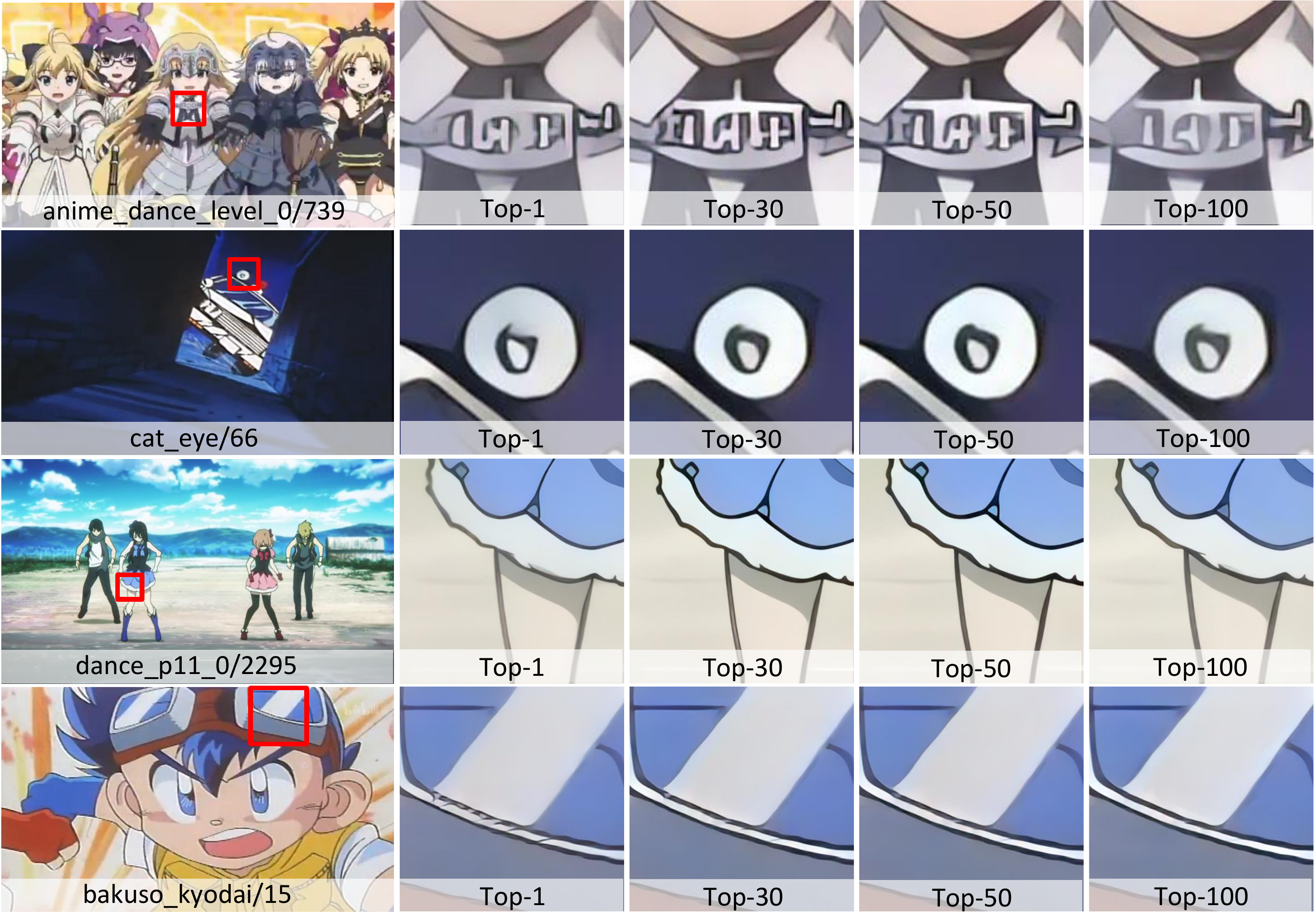}
    \caption{Ablation Study of the value K in stochastic Top-k VQ strategy. We choose K = 50 in our VQD-SR.}
    \label{fig:sup_topk}
\end{figure}

\begin{figure*}[]
    \centering
    \includegraphics[width=0.8\linewidth]{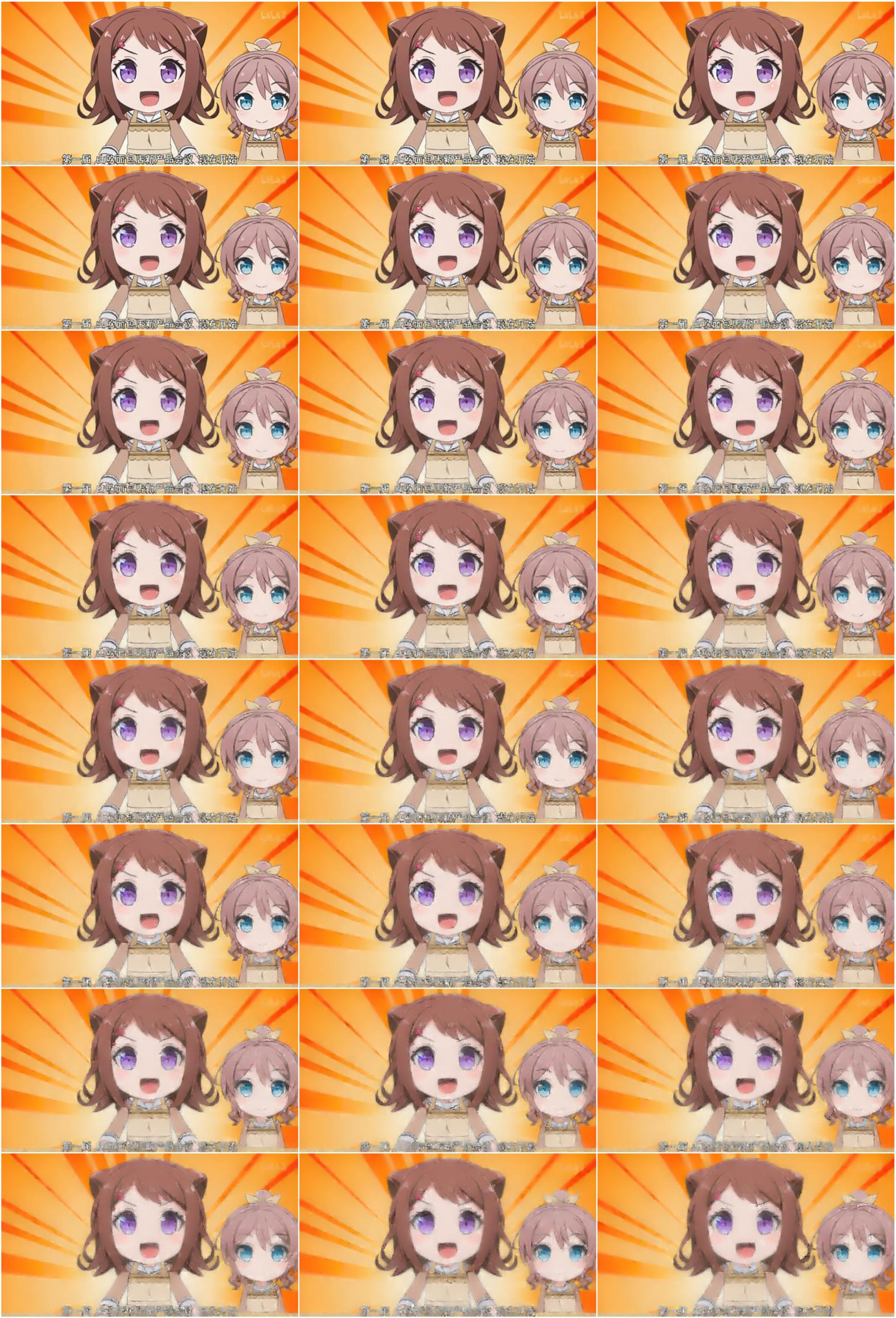}
    \caption{Examples of LR frames degraded by multi-scale VQGAN in multi-levels. From top to bottom, left to right, we show the degradation levels in ascending order. For the sake of clear comparisons, we show the results every 3 levels in the first 70 levels ($k = 1,4,7,...,70$) here. Input HR: AVC-Train$/bang\_dream\_p1\_0/00000049.png $}
    \label{fig:sup_degra1}
\end{figure*}

\begin{figure*}[]
    \centering
    \includegraphics[width=0.8\linewidth]{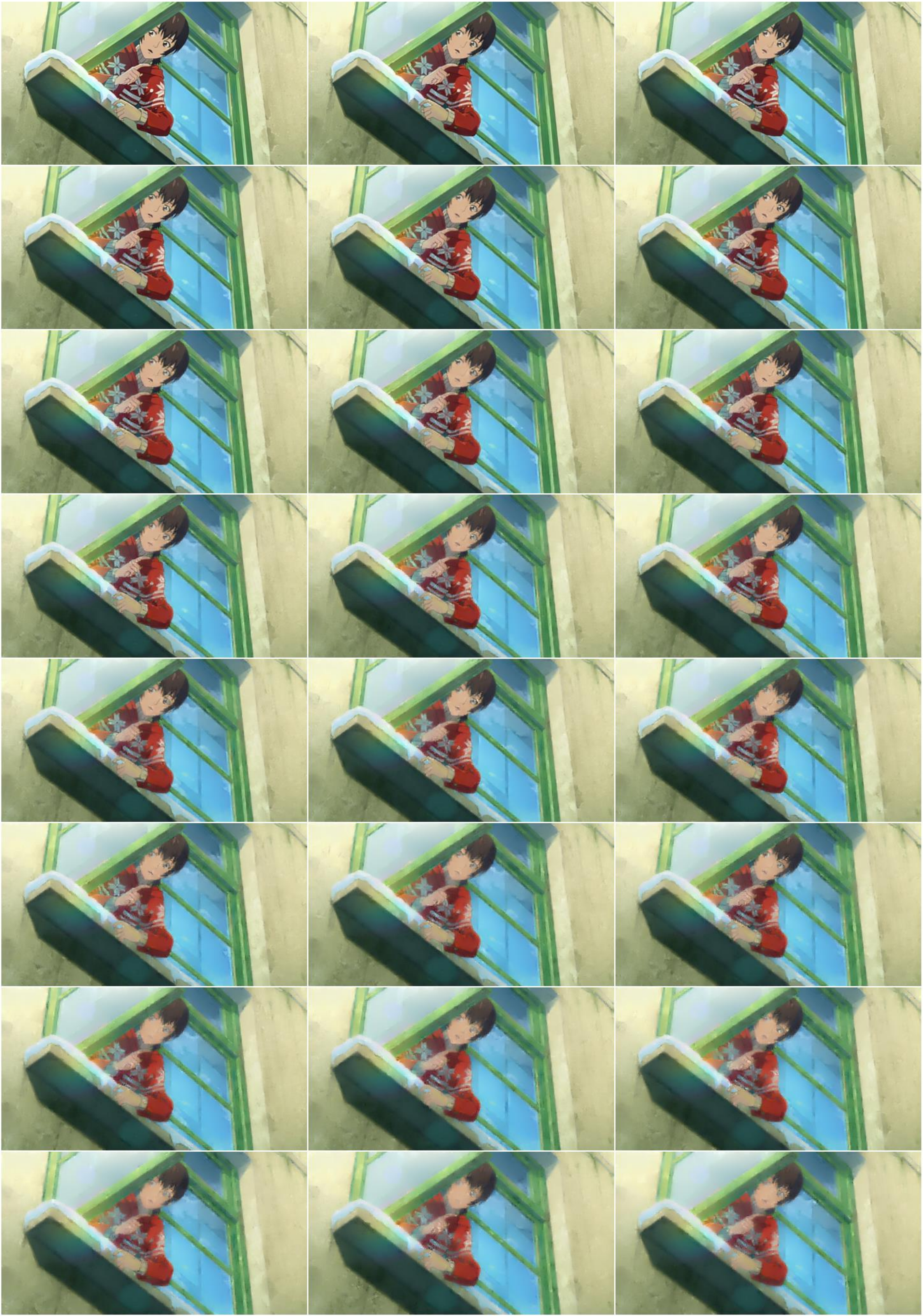}
    \caption{Examples of LR frames degraded by multi-scale VQGAN in multi-levels. From top to bottom, left to right, we show the degradation levels in ascending order. For the sake of clear comparisons, we show the results every 3 levels in the first 70 levels ($k = 1,4,7,...,70$) here. Input HR: AVC-Train$/b0034m9wleq.10005\_movie001\_0/00000049.png $}
    \label{fig:sup_degra2}
\end{figure*}

\noindent
\textbf{Stochastic Top-k VQ strategy.} Specifically, for each training clip in every iteration, we uniformly sample an integer $k$ from [1, $K$] as the degradation level, and utilize the $k_{th}$ nearest codebook entry to conduct the element-wise quantization on the encoded output in the bottom branch. A larger $k$ denotes a more severe degradation, and $K$ denotes the max degradation level in training. As shown in Tab.~\ref{tab:top_k_ablation} and Fig.~\ref{fig:sup_topk}, we compare the results of animation VSR when $K = \{1, 30, 50, 100\}$ on VQD-SR (base). When trained with a small $K$ (\eg, $K = 1$, also the nearest neighbor search), the animation VSR model has limited generalization ability due to the rigid degradation level in training. However, when $K$ is too large (\eg, $K = 100$), the degradations are so severe that contaminate the original image structures and disturb the training of VSR model. Based on the results, we finally choose $K = 50$ for our VQ degradation model, which leads to better results with sharper lines in smooth shapes. Referring to Fig.~\ref{fig:sup_degra1}-\ref{fig:sup_degra2}, we also show some visual examples of LR video frames degraded by multi-scale VQGAN in multi-levels with different $k$. 

\begin{table}[t]
    \centering
    \caption{Ablation Study of the value K in stochastic Top-k VQ strategy.}
    \vspace{1mm}
    \label{tab:top_k_ablation}
    \begin{tabular}{c||c|c|c|c}
    \Xhline{1.2pt}
                    & Top-1 & Top-30 & Top-50 & Top-100 \\
    \Xhline{0.4pt}
    MANIQA$\uparrow$& 0.3770 & 0.3846 & \textbf{0.3857} & 0.3756 \\
    \Xhline{1.2pt}
    \end{tabular}
\end{table}

\begin{table}[]
    \centering
    \caption{Results of different animation VSR methods in NIQE. `$*$' denotes fine-tune on animation dataset AVC-Train~\cite{AnimeSR}.}
    \vspace{1mm}
    \label{tab:niqe}
    \begin{tabular}{c || c |c |c}
    \Xhline{1.2pt}
                    & RealBasicVSR* & AnimeSR & VQD-SR \\
    \Xhline{0.4pt}
    NIQE$\downarrow$& 8.5358  & 8.7088 & \textbf{8.4737} \\
    \Xhline{1.2pt}
    \end{tabular}
    \vspace{-0.2cm}
\end{table}

\begin{table}[]
    \centering
    \caption{Evaluations with PSNR/SSIM on ATD test2k.}
    \vspace{1mm}
    \label{tab:atd_result}
    \begin{tabular}{c|| c c}
    \Xhline{1.2pt}
                    & AnimeSR & VQD-SR \\
    \Xhline{0.4pt}
    PSNR$\uparrow$& 35.49 &  \textbf{35.54} \\
    SSIM$\uparrow$& 0.9701 & \textbf{0.9702} \\
    \Xhline{1.2pt}
    \end{tabular}
    \vspace{-0.2cm}
\end{table}

\begin{table*}[]
    \centering
    \caption{Architecture of Multi-scale VQGAN. The proposed multi-scale VQGAN is composed of three parallel branches: (a) Top Branch, (b) Middle Branch, and (c) Bottom Branch. The downsample block is realized by a $3 \times 3$ convolution layer with stride 2. The upsample block is composed of a $3 \times 3$ convolution layer with stride 1 and a pixelshuffle layer. $C = 128$ is the number of base channel, $n_z = 256$ is the embedding dimension of VQ codebook. }
    \vspace{1mm}
    \label{tab:top_branch_arichtecture}
    \small
    \begin{tabular}{c||c}
    \Xhline{1.2pt}
     \multicolumn{2}{c}{\textbf{(a) Top Branch}} \\
    \Xhline{0.4pt}
       Encoder                                 &       Decoder \\
    \Xhline{0.4pt}
       $x\in \mathbb{R}^{H \times W \times 3}$ & $z_{q(t)}\in \mathbb{R}^{H/8 \times W/8 \times n_z}$\\
       Conv2D $ \rightarrow \mathbb{R}^{H \times W \times C} $ &      Conv2D $\rightarrow \mathbb{R}^{H/8 \times W/8 \times 4C}$ \\   
       Downsample Block, 2 $\times$ Residual Block{$\rightarrow x_b \in \mathbb{R}^{H/2 \times W/2 \times C}$} & Non-Local Block {$\rightarrow \mathbb{R}^{H/8 \times W/8 \times 4C}$}\\   
       Downsample Block, 2 $\times$ Residual Block $ \rightarrow x_m \in \mathbb{R}^{H/4 \times W/4 \times 2C}$ & 2 $\times$ Residual Block, Conv2D $\rightarrow \mathbb{R}^{H/8 \times W/8 \times 2C}$\\
       Downsample Block, 2 $\times$ Residual Block{$ \rightarrow \mathbb{R}^{H/8 \times W/8 \times 2C}$} & (+ $\hat{x}_m$), 2 $\times$ Residual Block, Upsample Block{$\rightarrow \mathbb{R}^{H/4 \times W/4 \times 2C}$} \\
       Conv2D, 2 $\times$ Residual Block{$\rightarrow \mathbb{R}^{H/8 \times W/8 \times 4C}$} & (+ $\hat{x}_b$), 2 $\times$ Residual Block, Upsample Block{$\rightarrow \mathbb{R}^{H/2 \times W/2 \times C}$}\\
       Non-Local Block{$\rightarrow \mathbb{R}^{H/8 \times W/8 \times 4C}$} & 2 $\times$ Residual Block, Upsample Block{$\rightarrow \mathbb{R}^{H \times W \times C}$}\\
       GroupNorm, Conv2D {$\rightarrow z_t \in \mathbb{R}^{H/8 \times W/8 \times n_z}$} &  2 $\times$ Residual Block {$\rightarrow \hat{x} \in \mathbb{R}^{H \times W \times 3}$}\\
    \Xhline{1.2pt}
       
    \multicolumn{2}{c}{\textbf{(b) Middle Branch}} \\
    \Xhline{0.4pt}
    Encoder                                 &       Decoder \\
    \Xhline{0.4pt}
       $x_m\in \mathbb{R}^{H/4 \times W/4 \times 2C}$ & $z_{q(m)}\in \mathbb{R}^{H/4 \times W/4 \times n_z}$ \\  
       Conv2D, 2 $\times$ Residual Block{$\rightarrow \mathbb{R}^{H/4 \times W/4 \times 2C}$} &  Conv2D$\rightarrow \mathbb{R}^{H/4 \times W/4 \times 4C}$ \\
       Conv2D, 2 $\times$ Residual Block{$\rightarrow \mathbb{R}^{H/4 \times W/4 \times 4C}$} & 2 $\times$ Residual Block, Conv2D $\rightarrow \mathbb{R}^{H/4 \times W/4 \times 2C}$ \\
       GroupNorm, Conv2D {$\rightarrow z_m \in \mathbb{R}^{H/4 \times W/4 \times n_z}$} &  2 $\times$ Residual Block, Conv2D $\rightarrow \hat{x}_m \in \mathbb{R}^{H/4 \times W/4 \times 2C}$\\

    \Xhline{1.2pt}
    \multicolumn{2}{c}{\textbf{(c) Bottom Branch}} \\
    \Xhline{0.4pt}
    Encoder                                 &       Decoder \\
    \Xhline{0.4pt}
      $x_b\in \mathbb{R}^{H/2 \times W/2 \times C}$ & $z_{q(b)}\in \mathbb{R}^{H/2 \times W/2 \times n_z}$ \\  
       Conv2D, 2 $\times$ Residual Block{$\rightarrow \mathbb{R}^{H/2 \times W/2 \times 2C}$} &  Conv2D$\rightarrow \mathbb{R}^{H/2 \times W/2 \times 4C}$ \\
       Conv2D, 2 $\times$ Residual Block{$\rightarrow \mathbb{R}^{H/2 \times W/2 \times 2C}$} & 2 $\times$ Residual Block, Conv2D $\rightarrow \mathbb{R}^{H/2 \times W/2 \times 2C}$ \\
       Conv2D, 2 $\times$ Residual Block{$\rightarrow \mathbb{R}^{H/2 \times W/2 \times 4C}$} & 2 $\times$ Residual Block, Conv2D $\rightarrow \mathbb{R}^{H/2 \times W/2 \times 2C}$ \\
       GroupNorm, Conv2D {$\rightarrow z_b \in \mathbb{R}^{H/2 \times W/2 \times n_z}$} &  2 $\times$ Residual Block, Conv2D $\rightarrow \hat{x}_b \in \mathbb{R}^{H/2 \times W/2 \times C}$\\
    
    \Xhline{1.2pt}
    \end{tabular}
\end{table*}

\subsection{Video Super-Resolution Model}
\label{subsec:vsr}
\begin{figure}[]
    \centering
    \includegraphics[width=0.9\linewidth]{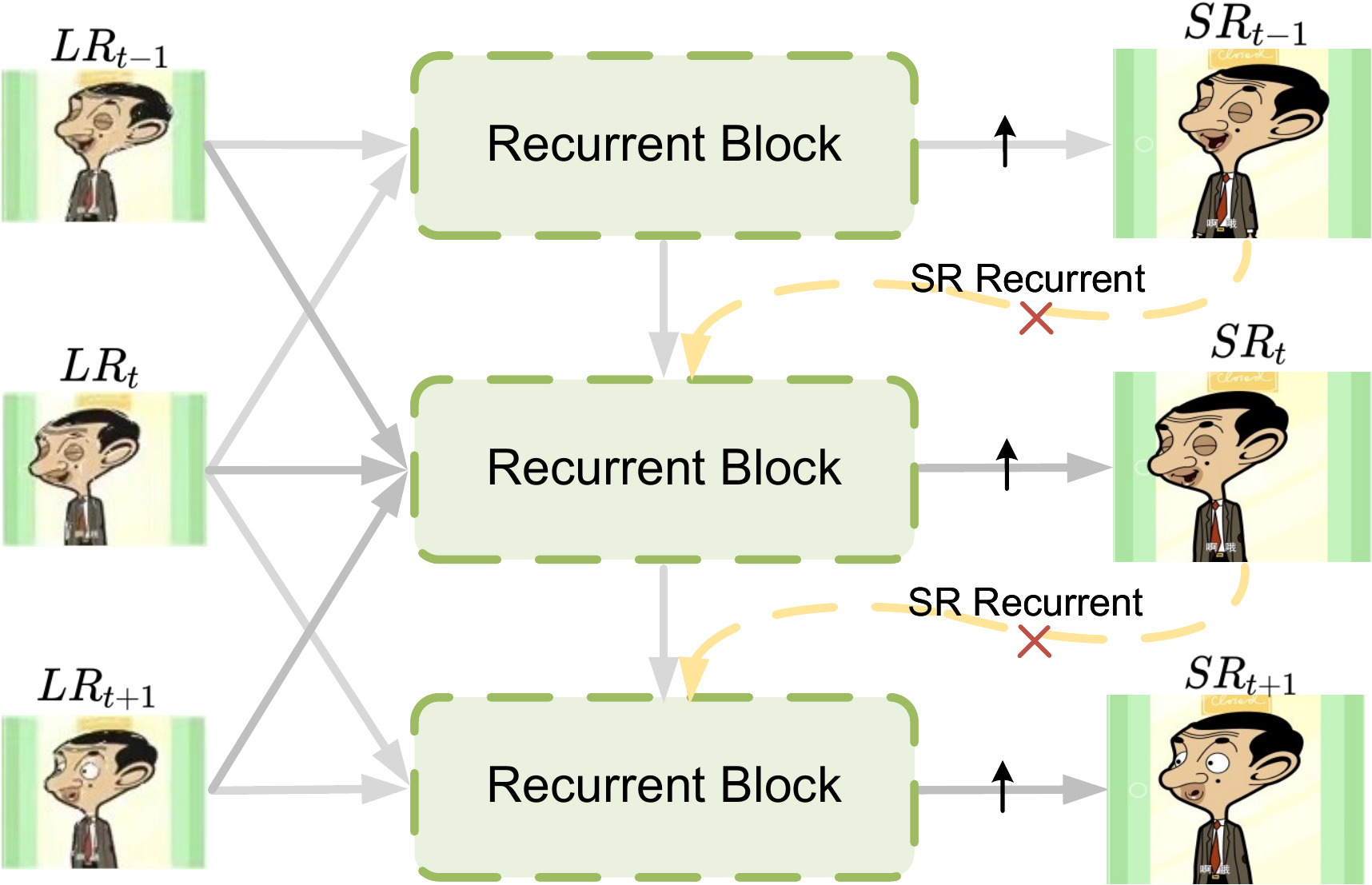}
    \caption{Architecture of VSR model. We follow the VSR model in AnimeSR~\cite{AnimeSR} but remove the SR feedback in the recurrent block for stable training.}
    \label{fig:sup_vsr_model}
\end{figure}

\begin{figure}[]
    \centering
    \includegraphics[width=\linewidth]{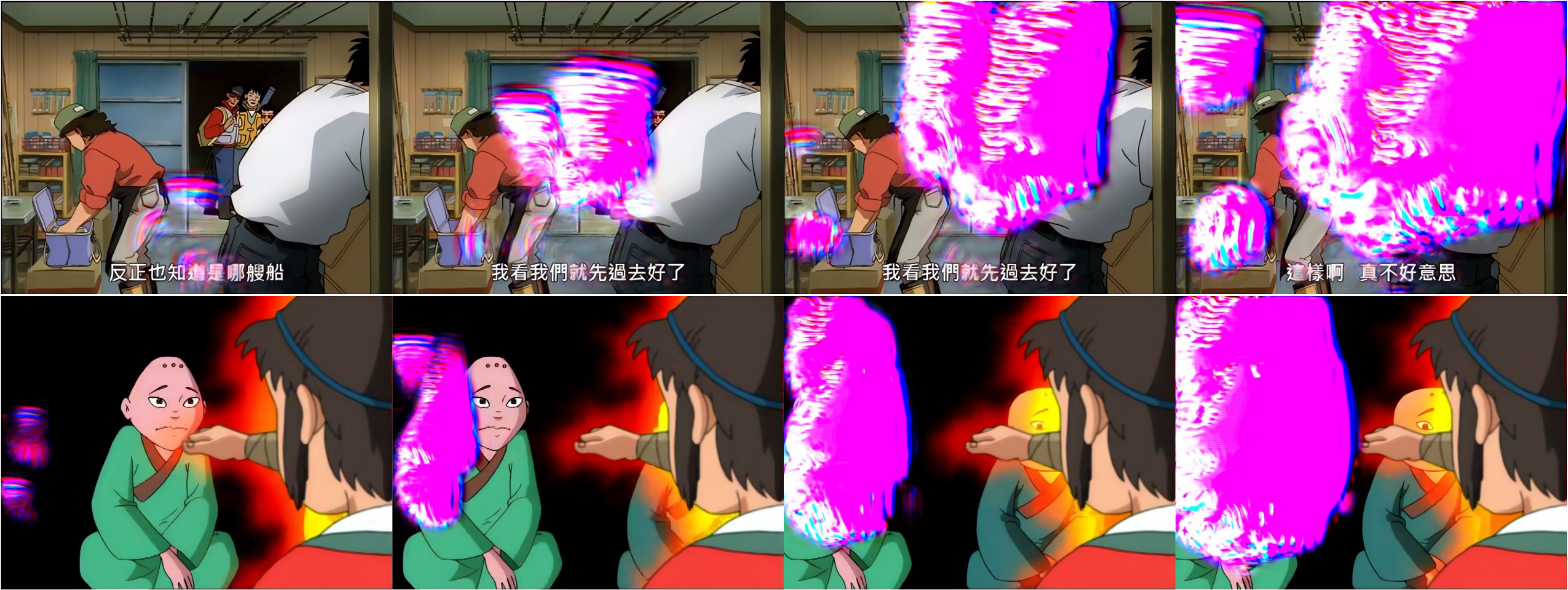}
    \caption{Too many recurrent features cause unstable training leading to corrupt animation VSR models.}
    \label{fig:sup_collapse}
\end{figure}

We follow the VSR model in AnimeSR~\cite{AnimeSR} because of its efficiency but remove the SR feedback in the recurrent block as shown in Fig.~\ref{fig:sup_vsr_model}. Different from natural domain videos, the continuities between animation video frames are relatively poor which causes difficulties for explicit alignment modules and further impacts the final VSR results. Thus, the explicit alignment module is left out in the architecture of animation VSR model, where the misaligned recurrent features are directly adopted, which also greatly shrinks the computation costs. However, as is studied by Chan \etal~\cite{RealBasicVSR} that although long-term information is beneficial for VSR, it may suffer from error accumulation during propagation. For VSR models without explicit alignment modules, this problem could be more severe. As shown in Fig.~\ref{fig:sup_collapse}, we find that too many misaligned recurrent features make the animation VSR model~\cite{AnimeSR} susceptible to error accumulation, and sometimes cause a collapse of GAN loss in training, leading to corrupt VSR models. Based on this observation, we remove the SR feedback in the recurrent block to ensure more stable training. We train the VSR model on eight NVIDIA 32G V100 GPUs.

\subsection{Experiment Details}
\label{subsec:exper}
\noindent
\textbf{Evaluations with MANIQA.} We test the super-resolution methods on AVC-RealLQ~\cite{AnimeSR}, which is a real-world animation video dataset containing 46 low-quality clips with 100 frames per clip. As the lack of ground truths for testing, we follow AnimeSR~\cite{AnimeSR} and adopt the no-reference image quality assessment (NR-IQA) metric MANIQA~\cite{yang2022maniqa} to evaluate the final SR results in the main paper. MANIQA is designed for rating GAN-based distorted images, which is suitable for evaluating GAN-based image restoration algorithms. Following \cite{yang2022maniqa} and \cite{AnimeSR}, we measure MANIQA every 10 frames in each video. For each testing frame, we randomly crop $224 \times 224$ sized images 20 times and calculate the average score. We run the evaluation process 3 times and report the mean metrics for all of the reported results.

\noindent
\textbf{Evaluations with NIQE.} 
We show the evaluations of different animation VSR methods with NIQE~\cite{mittal2012making} on AVC-RealLQ in Tab.~\ref{tab:niqe}. NIQE is a hand-crafted feature-based method that has less ability to measure the diverse distortions in the real world and especially some GAN-based distortions caused by image restoration algorithms. \cite{AnimeSR} also has mentioned that NIQE lacks consistent with the perceptual visual quality. However, our VQD-SR still outperforms the other two SOTA animation VSR methods in NIQE.

\begin{figure}[]
    \centering
    \includegraphics[width=\linewidth]{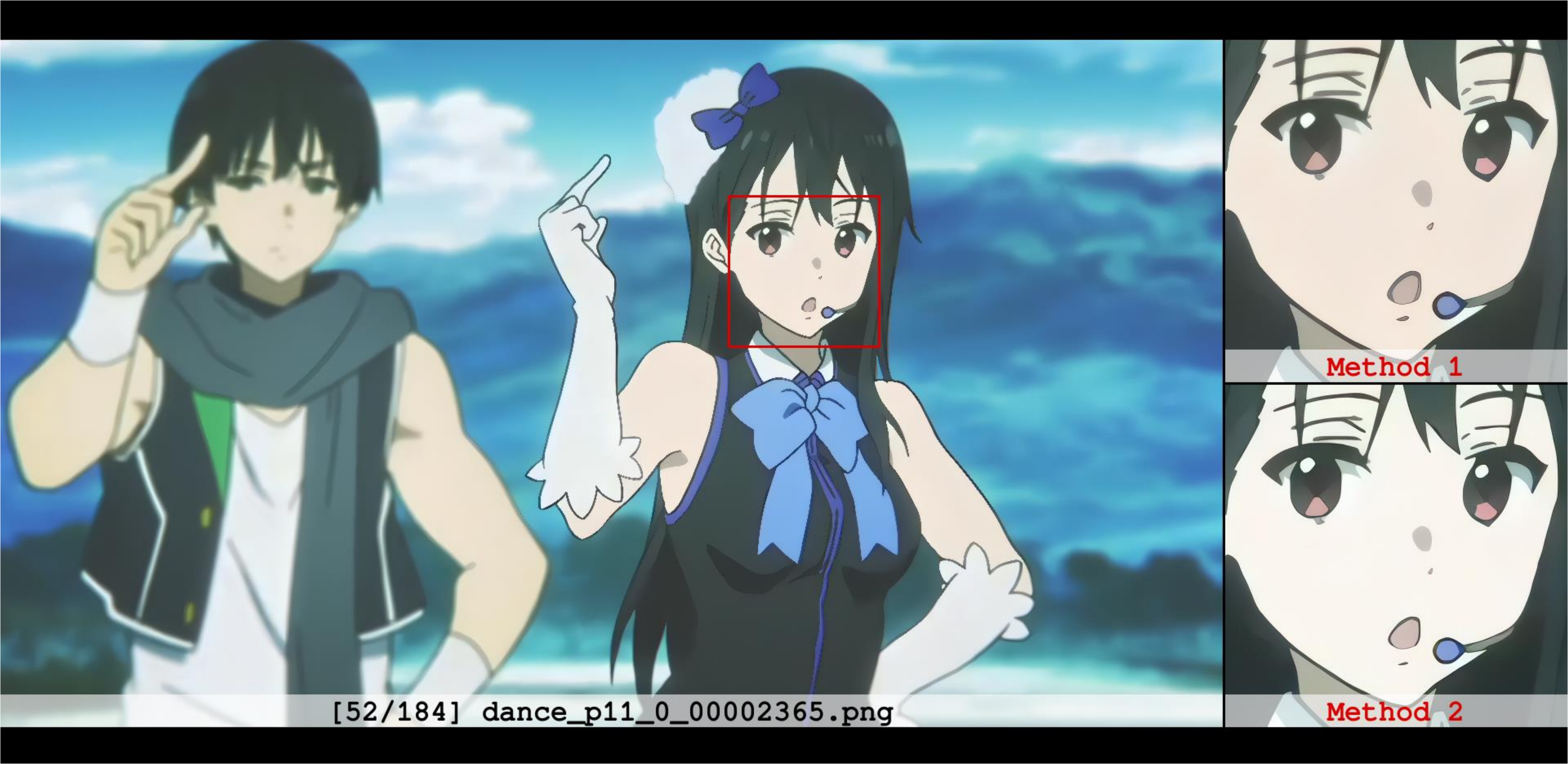}
    \caption{UI of the A-B test. The main window shows the complete image with the current testing progress and frame name beneath. Two sub-windows on the right display the zoom-in details of two methods where the mouse hovers, and the size of the viewport can also be adjusted by scrolling the mouse wheel. Users are asked to select the one with higher visual quality by pressing `1' or `2' on the keyboard.}
    \label{fig:sup_ui}
\end{figure}

\noindent
\textbf{Evaluations on ATD test2k}. ATD-12K~\cite{siyao2021deep} is a large-scale HR animation triplet dataset, which comprises 12,000 triplets. As only HR images are provided in ATD, we follow the conventional SR setting that reports PSNR/SSIM of AnimeSR and ours on $4\times$ bicubic (BI) downsampled images in Tab.~\ref{tab:atd_result}. The results demonstrate the advantage of our method under the ideal BI setting. However, the setting of AVC-RealLQ in the main paper is a more challenging but practical scenario for real applications of animation VSR. 

\noindent
\textbf{User Study on Visual Quality.} We conduct A-B tests to further compare the visual quality of VQD-SR with other six SOTA methods (BasicVSR~\cite{chan2021basicvsr}, PDM~\cite{luo2022learning}, Real-ESRGAN~\cite{wang2021real}, BSRGAN~\cite{zhang2021designing}, RealBasicVSR~\cite{RealBasicVSR}, AnimeSR~\cite{AnimeSR}) successively. For each comparison with one of the six methods, there are 20 subjects involved in the tests on the SR results of AVC-RealLQ~\cite{AnimeSR} with 46 animation video clips. Considering the subtle qualitative differences between frames, we uniformly sample 4 frames for testing in each clip. The user interface for the A-B test, as shown in Fig.~\ref{fig:sup_ui}, provides the users with two images in random order which include one VQD-SR frame and one frame from the other method. Users are asked to select one with higher visual quality. As the resolutions of SR results are too large (\eg, 5760 $\times$ 4320) to fit the screen and display the details at the same time if only providing the complete images, we further show two side-by-side zoom-in windows controlled by mouse with adjustable positions and sizes. The final results are the percentages of votes which prefer VQD-SR to other methods. 

\section{Statistics and Samples of RAL}
\label{sec:ral_sta}
\begin{figure}[]
    \centering
    \includegraphics[width=\linewidth]{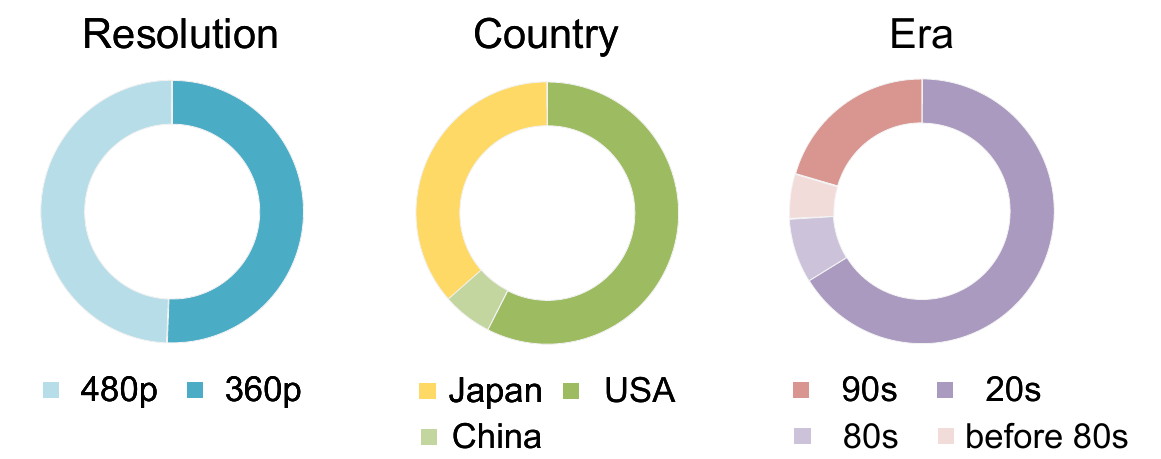}
    \caption{Statistics of RAL.}
    \label{fig:sup_ral_sta}
\end{figure}

\begin{figure}[]
    \centering
    \includegraphics[width=\linewidth]{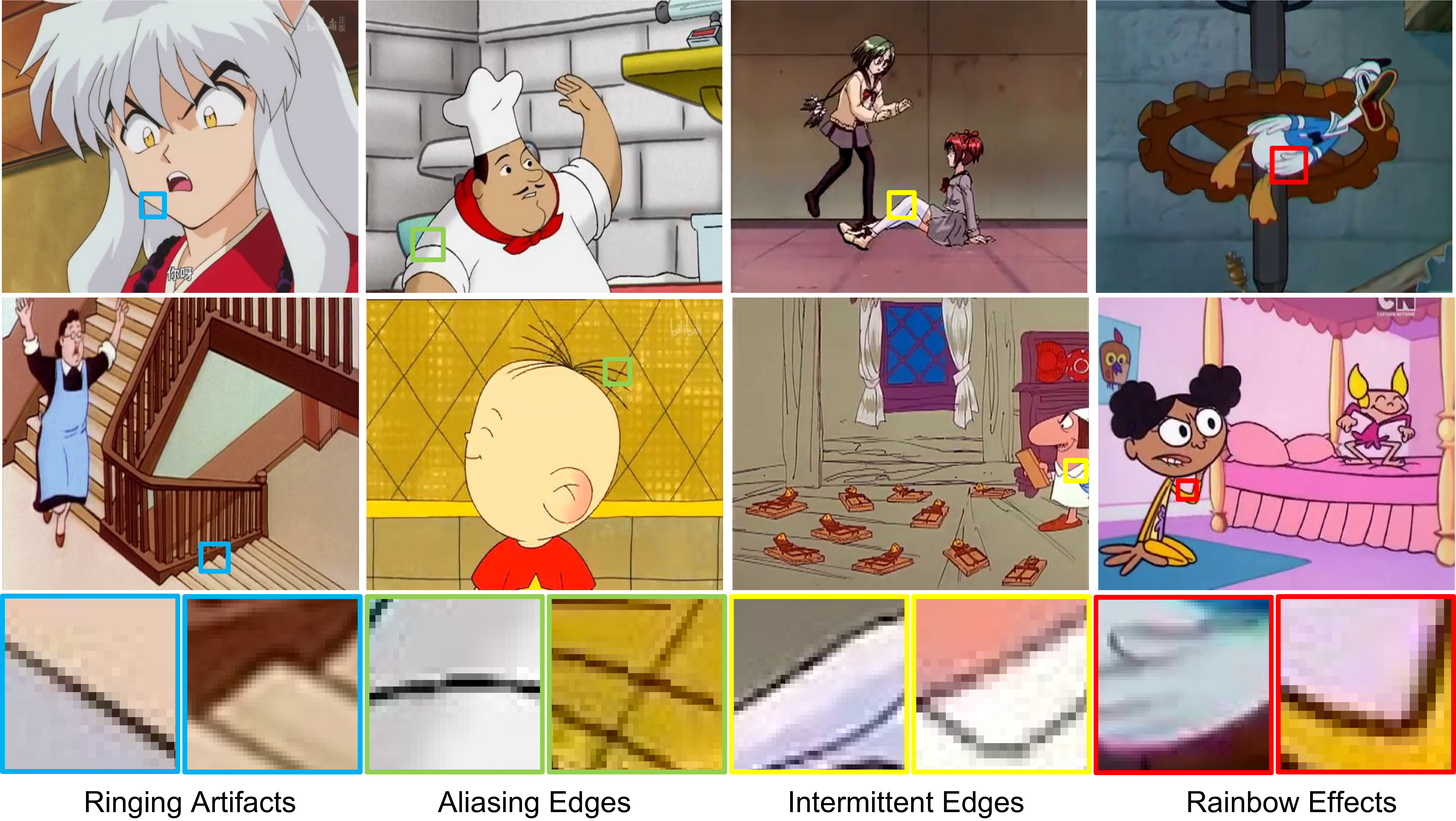}
    \caption{Samples of RAL and typical degradation phenomena in real-world LR animation videos.}
    \label{fig:sup_ral_degra}
\end{figure}

Our \textbf{R}eal \textbf{A}nimation \textbf{L}ow-quality (RAL) video dataset contains over 10K LR frames extracted from 441 real-world low-quality animation videos and contains rich real-world degradations in animation domain. The statistics of RAL is shown in Fig.~\ref{fig:sup_ral_sta}. In Fig.~\ref{fig:sup_ral_degra}, we also show some representative samples and typical real-world degradations in RAL.

\section{More Qualitative Comparisons}
\label{sec:more_results}
In this section, we show more qualitative results to verify the effectiveness of our proposed methods. 

\begin{figure*}
    \centering
    \includegraphics[width=0.9\linewidth]{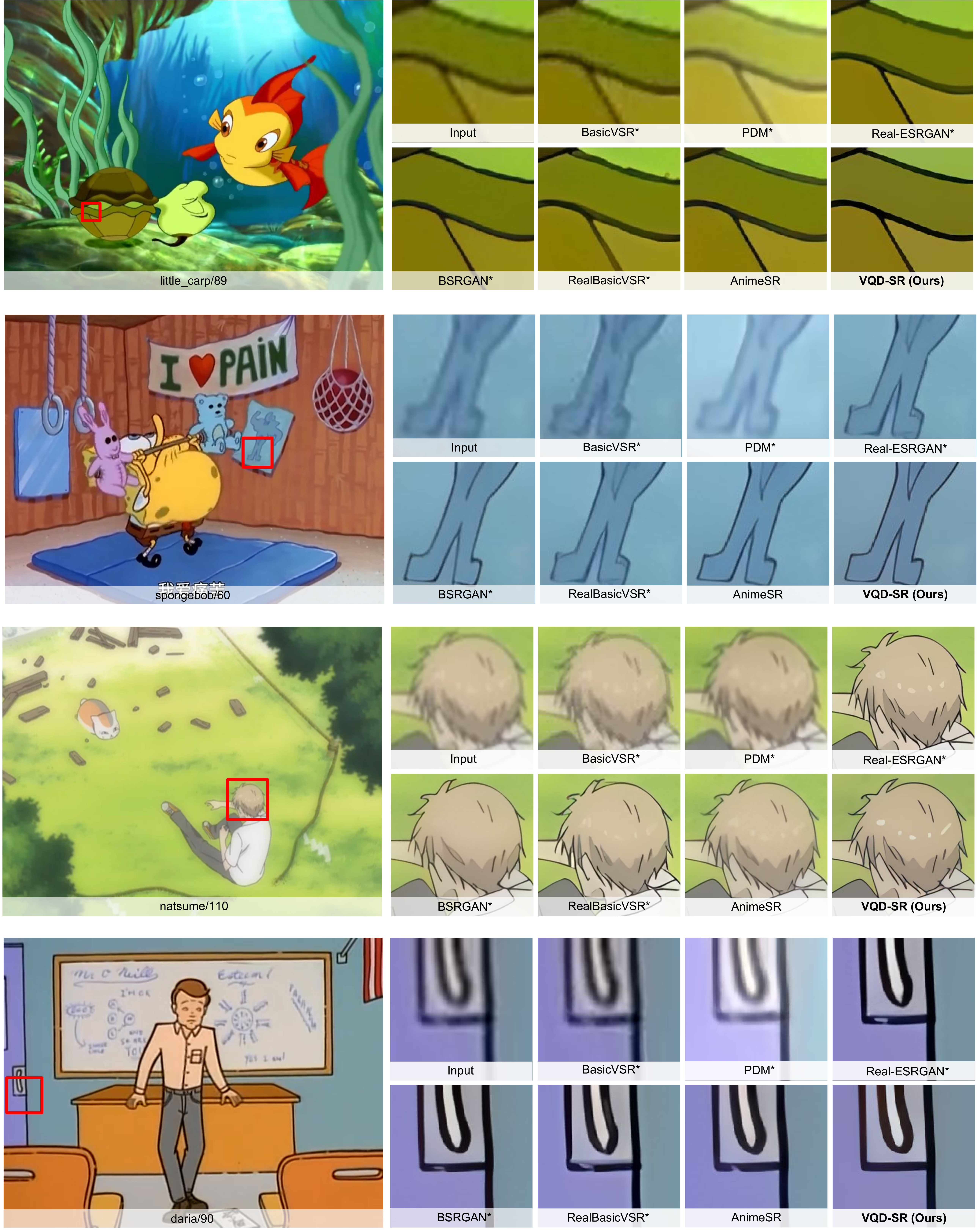}
    \caption{Qualitative comparisons with SOTA methods. `$*$' denotes fine-tune on animation dataset AVC-Train~\cite{AnimeSR}. Our VQD-SR is capable to recover visually natural and sharper lines with fewer artifacts.}
    \label{fig:sup_main_1}
\end{figure*}

\begin{figure*}
    \centering
    \includegraphics[width=0.9\linewidth]{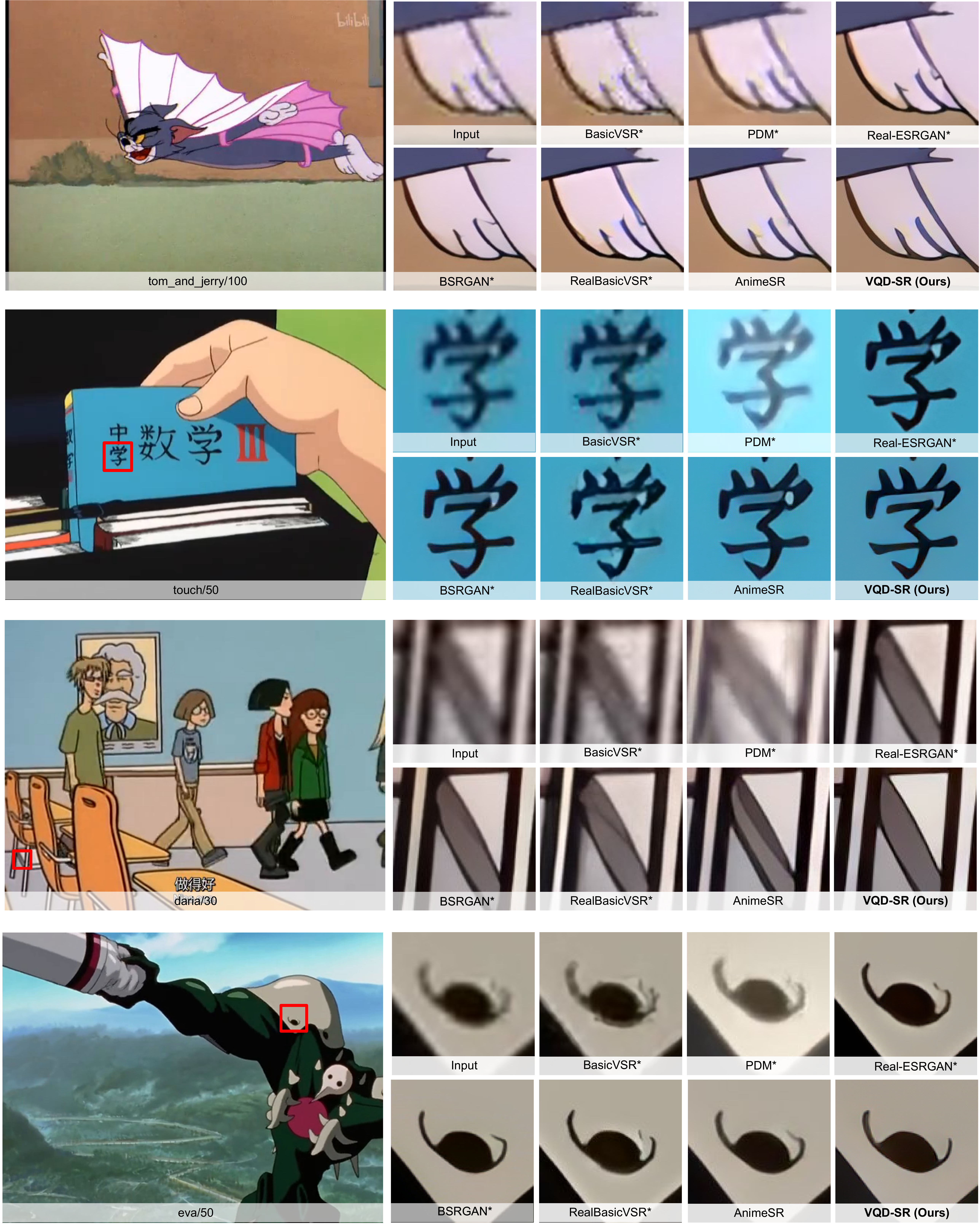}
    \caption{Qualitative comparisons with SOTA methods. `$*$' denotes fine-tune on animation dataset AVC-Train~\cite{AnimeSR}. Our VQD-SR is capable to restore clear details.}
    \label{fig:sup_main_2}
\end{figure*}

\begin{figure*}
    \centering
    \includegraphics[width=0.9\linewidth]{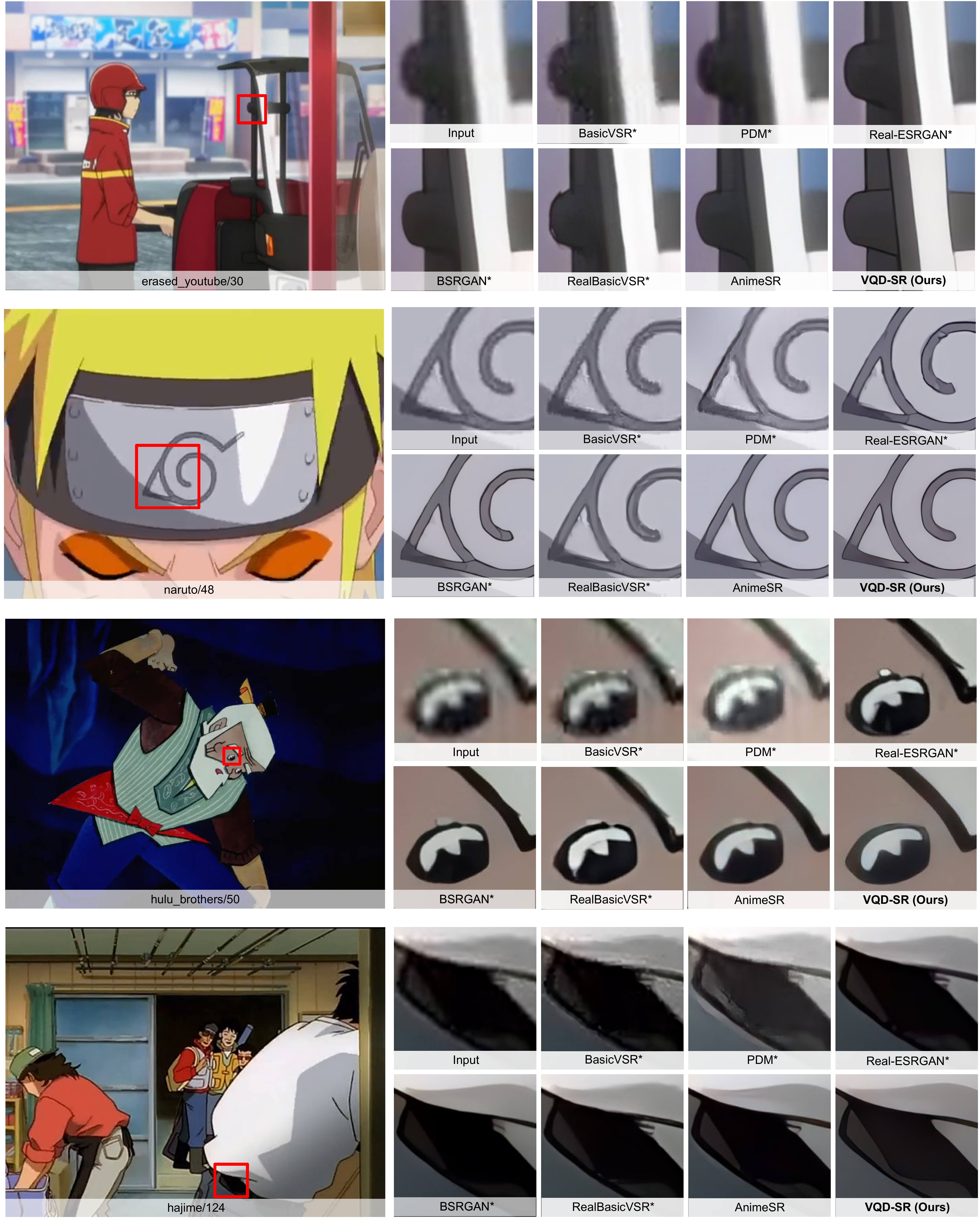}
    \caption{Qualitative comparisons with SOTA methods. `$*$' denotes fine-tune on animation dataset AVC-Train~\cite{AnimeSR}. Our VQD-SR is capable to recover visually natural and sharper lines with fewer artifacts.}
    \label{fig:sup_main_3}
\end{figure*}

\begin{figure*}
    \centering
    \includegraphics[width=\linewidth]{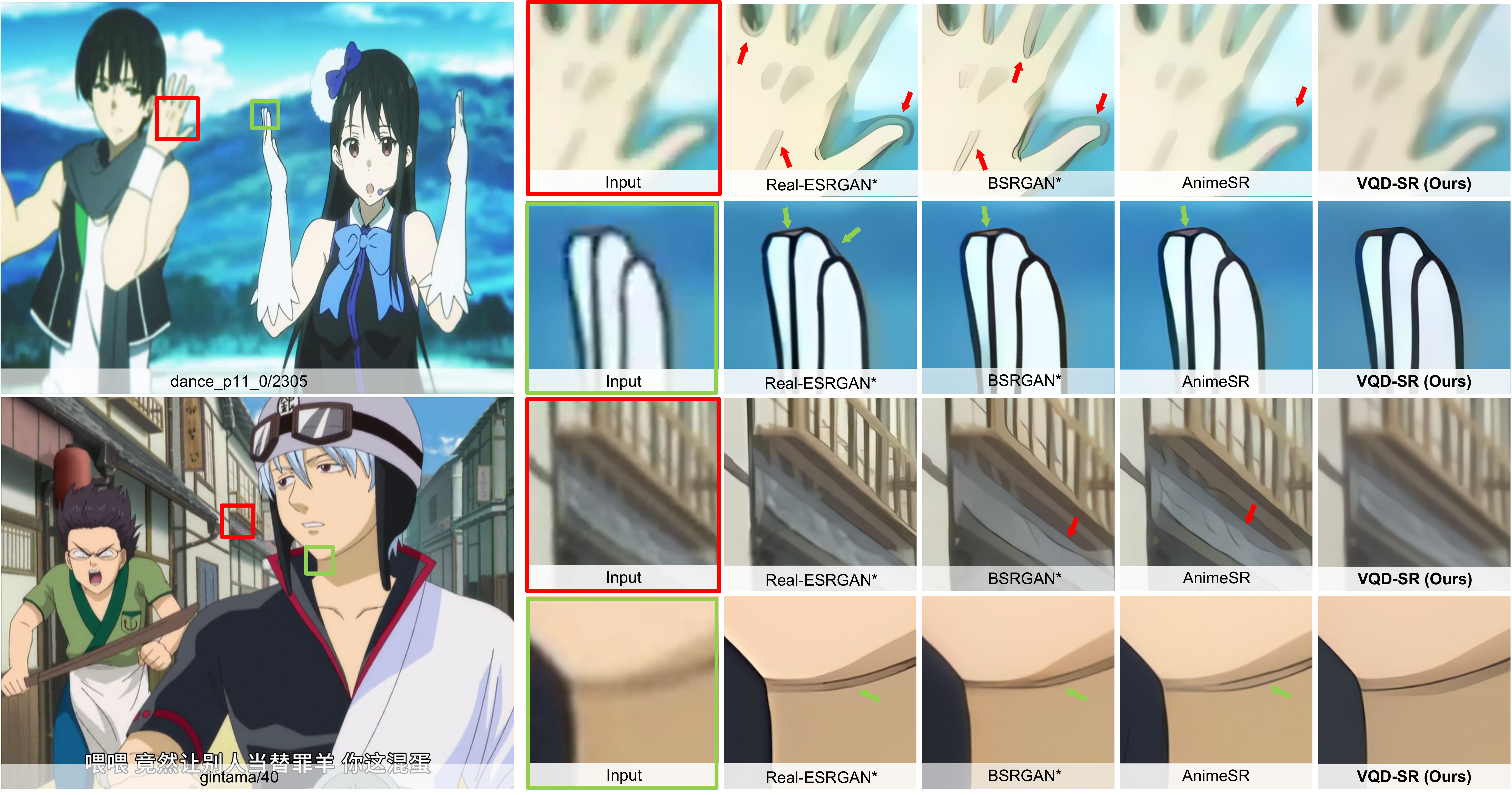}
    \caption{Qualitative comparisons with SOTA methods. `$*$' denotes fine-tune on animation dataset AVC-Train~\cite{AnimeSR}. Our VQD-SR is capable to handle some intended scenarios (\eg, the out-focus background blur) with fewer over-sharp artifacts. The \textcolor{red}{red} crops indicate objects in the out-focus background which should be naturally smooth and the \textcolor{green}{green} crops indicate the foreground objects which should be clear and sharp.}
    \label{fig:sup_blur_effect}
\end{figure*}

\noindent
\textbf{Animation VSR Results.} In Fig.~\ref{fig:sup_main_1} - \ref{fig:sup_blur_effect}, we compare our VQD-SR with six SOTA SR methods. The first crop shown in each case is the bicubic $4\times$ upscaled original input for reference, as the absence of ground truths in real scenarios. Our VQD-SR is capable to recover visually natural and sharper lines (Fig.~\ref{fig:sup_main_1}, Fig.\ref{fig:sup_main_3}) with fewer artifacts, restore clear details (Fig.~\ref{fig:sup_main_2}), handle some intended scenarios (\eg, the out-focus background blur) with fewer over-sharp artifacts (Fig.~\ref{fig:sup_blur_effect}). 

\begin{figure}[h]
    \centering
    \includegraphics[width=\linewidth]{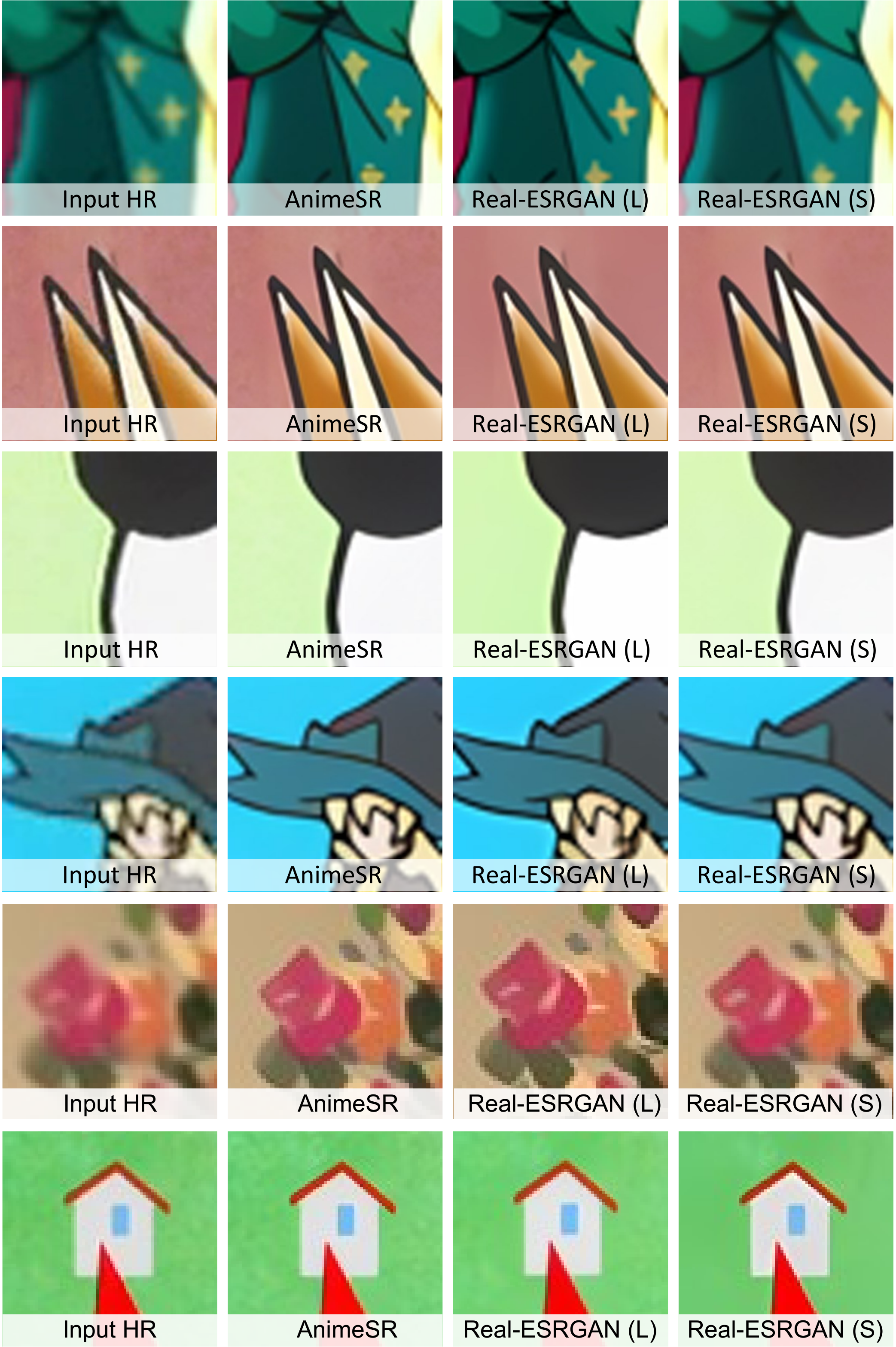}
    \caption{Visual comparison of different SR models for HR-SR enhancement in animation domain.}
    \label{fig:sup_hr_ani}
\end{figure}

\begin{figure*}
    \centering
    \includegraphics[width=0.7\linewidth]{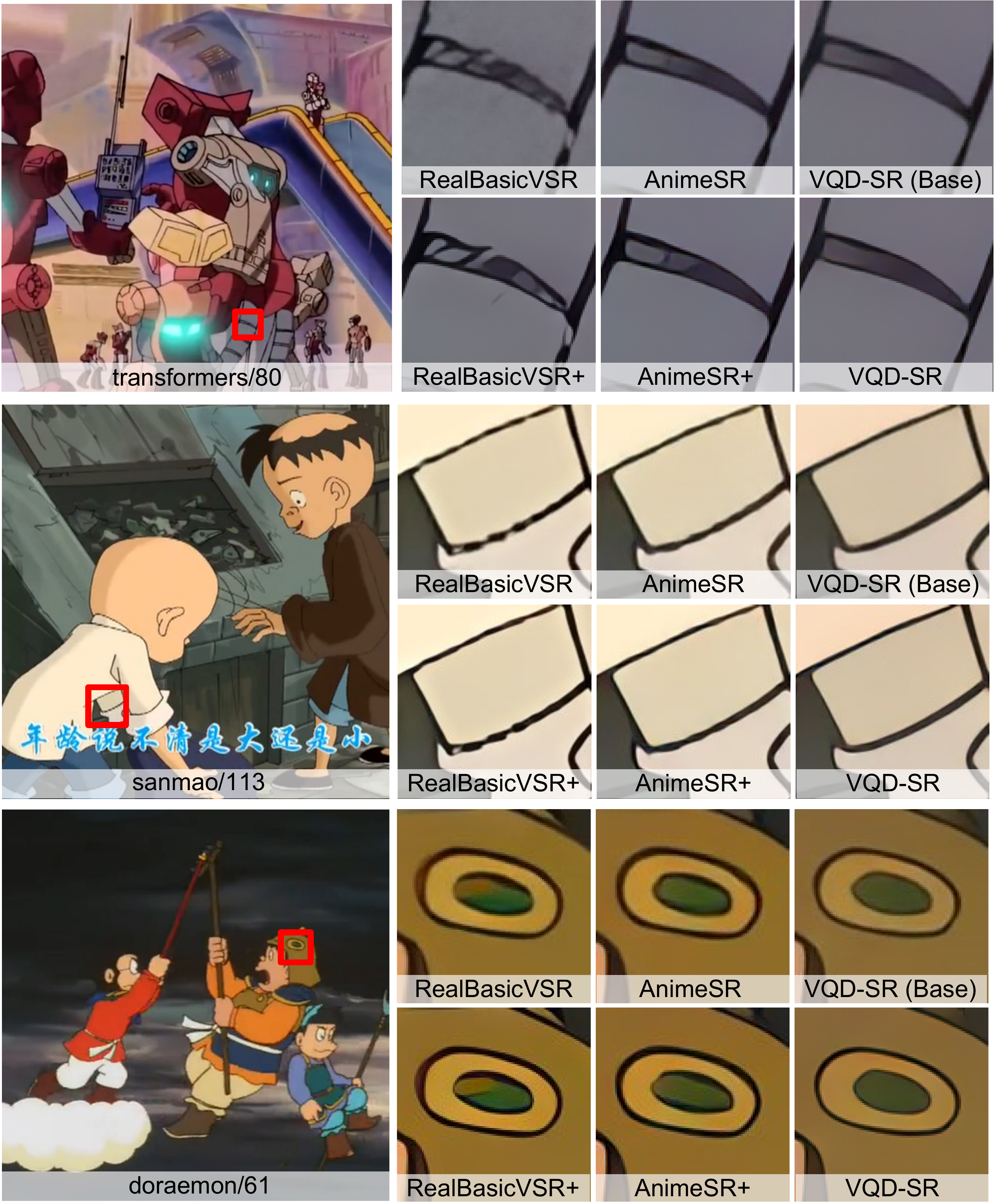}
    \caption{Ablation study of HR-SR enhancement for different VSR methods.}
    \label{fig:sup_hr_abl}
\end{figure*}

\noindent
\textbf{HR-SR Enhancement.} In Fig.~\ref{fig:sup_hr_ani}, we show the enhanced HR animation video frames with different SR models. Our HR-SR enhancement strategy could alleviate the compression artifacts and sharpen the edges without contaminating the original details in animation HR frames. As shown in Fig.~\ref{fig:sup_hr_abl}, the proposed HR-SR strategy is valid to improve the results of animation VSR, regardless of the specific VSR model, with the help of more effective ground truths for training.

\begin{figure}
    \centering
    \includegraphics[width=\linewidth]{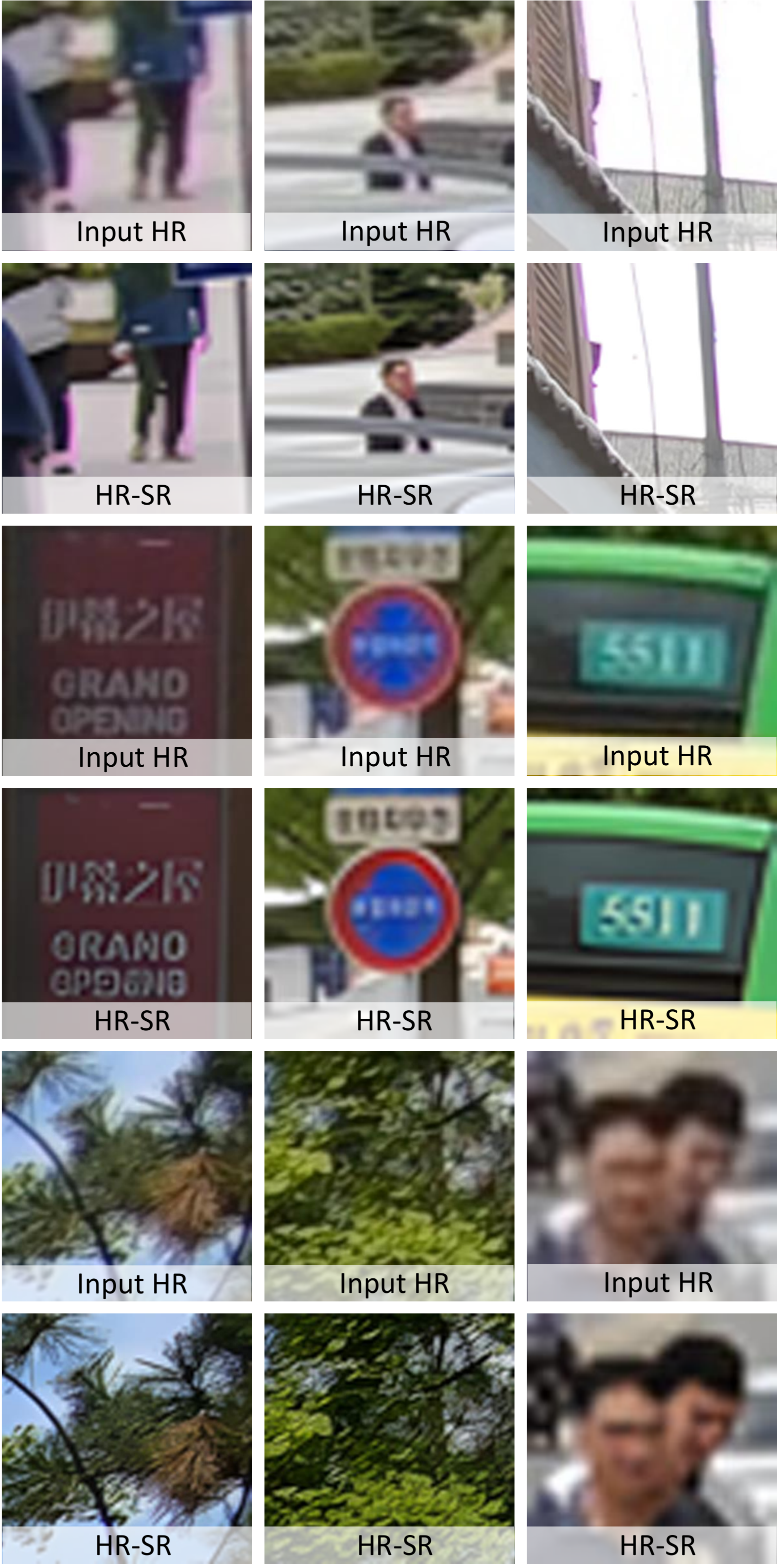}
    \caption{Extend HR-SR enhancement strategy to natural videos. Complex textures and irregular illumination conditions lead to amplified illumination artifacts (row \textcolor{red}{1}), contaminated details (row \textcolor{red}{2}), and over-sharp textures (row \textcolor{red}{3}).}
    \label{fig:sup_hr_natur}
\end{figure}

We further extend the HR-SR enhancement strategy from animation videos to natural videos (REDS~\cite{nah2019ntire}) in Fig.~\ref{fig:sup_hr_natur}. Because of the complex textures and irregular illumination conditions, directly adopting HR-SR enhancement strategy to natural videos would cause amplified illumination artifacts (row \textcolor{red}{1}), contaminated details (row \textcolor{red}{2}), and over-sharp textures (row \textcolor{red}{3}), leading to unappealing results.  

\end{appendix}
\end{document}